\documentclass[11pt,a4paper]{article}
\usepackage[hyperref]{acl2020}
\usepackage{times}
\usepackage{latexsym}
\usepackage[pdftex]{graphicx}

\usepackage{dsfont}
\usepackage{bbm}
\usepackage{mathbbol}
\usepackage{xcolor}
\usepackage{MnSymbol}
\usepackage{amsmath,amsfonts}
\usepackage[capitalize]{cleveref}
\usepackage{url}
\usepackage{color,soul}
\usepackage{bm}
\usepackage{multirow}
\usepackage[font=small]{caption}
\usepackage{subcaption}
\usepackage[noend]{algpseudocode}
\usepackage{afterpage}
\usepackage{tabularx}
\usepackage{booktabs}
\usepackage{arydshln}
\usepackage{url}
\usepackage[shortlabels]{enumitem}
\usepackage{bigstrut}
\usepackage{array}
\usepackage{microtype}
\usepackage{pifont}
\usepackage{relsize}
\usepackage{graphicx}       
\usepackage{cleveref}
\usepackage{todonotes}
\usepackage{xcolor}
\usepackage{adjustbox}
\definecolor{darkgreen}{rgb}{0,0.5,0}

\crefformat{section}{\S#2#1#3} 
\crefformat{subsection}{\S#2#1#3}
\crefformat{subsubsection}{\S#2#1#3}

\graphicspath{{img/}}

\usepackage{floatrow}
\setkeys{Gin}{width=\linewidth}

\newcommand{\norm}[1]{\left\lVert#1\right\rVert}
\DeclareMathOperator*{\argmin}{arg\,min} 

\aclfinalcopy 
\def\aclpaperid{2371} 

\title{Kin{GDOM}: {K}nowledge-{G}uided {DOM}ain adaptation\\ for sentiment analysis}

\author{Deepanway Ghosal\(^\dagger\) , \ Devamanyu Hazarika\(^\Phi\) , Abhinaba Roy\(^\oslash\),\\
\textbf{Navonil Majumder\(^\dagger\),
 \ Rada Mihalcea\(^\sqcap\) , \ Soujanya Poria\(^\dagger\) }\\ \\
\(^\dagger\)Singapore University of Technology and Design, Singapore\\
\(^\Phi\)National University of Singapore, Singapore\\
\(^\oslash\)Nanyang Technological University, Singapore  \\
\(^\sqcap\)University of Michigan, USA\\
\fontsize{10}{12}\texttt{\{deepanway\_ghosal@mymail, sporia@, navonil\_majumder@\}.sutd.edu.sg},\\ 
\fontsize{10}{12}\texttt{hazarika@comp.nus.edu.sg}, \texttt{abhinaba.roy@ntu.edu.sg}, \texttt{mihalcea@umich.edu}\\
}

\date{}

\begin{document}
\maketitle
\begin{abstract}
Cross-domain sentiment analysis has received significant attention in recent years, prompted by the need to combat the domain gap between different applications that make use of sentiment analysis. In this paper, we take a novel perspective on this task by exploring the role of external commonsense knowledge. We introduce a new framework, \textit{KinGDOM}, which utilizes the ConceptNet knowledge graph to enrich the semantics of a document by providing both domain-specific and domain-general background concepts. These concepts are learned by training a graph convolutional autoencoder that leverages inter-domain concepts in a domain-invariant manner. Conditioning a popular domain-adversarial baseline method with these learned concepts helps improve its performance over state-of-the-art approaches, demonstrating the efficacy of our proposed framework.
\end{abstract}

\section{Introduction} \label{section:intro}

Sentiment Analysis (SA) is a popular NLP task used in many applications ~\cite{DBLP:journals/widm/ZhangWL18}. Current models trained for this task, however, cannot be reliably deployed due to the distributional mismatch between the training and evaluation domains~\cite{DBLP:journals/jair/DaumeM06}. \textit{Domain adaptation}, a case of transductive transfer learning, is a widely studied field of research that can be effectively used to tackle this problem~\cite{DBLP:journals/corr/abs-1812-02849}.

Research in the field of cross-domain SA has proposed diverse approaches, which include learning domain-specific sentiment words/lexicons~\cite{DBLP:conf/acl/SarmaLS18,hamilton-etal-2016-diachronic}, co-occurrence based learning~\cite{DBLP:conf/acl/BlitzerDP07}, domain-adversarial learning~\cite{JMLR:v17:15-239}, among others.
In this work, we adopt the domain-adversarial framework and attempt to improve it further by infusing commonsense knowledge using ConceptNet -- a large-scale knowledge graph~\cite{speer2017conceptnet}.

\begin{figure}[t]
    \centering
    \includegraphics[width=\linewidth]{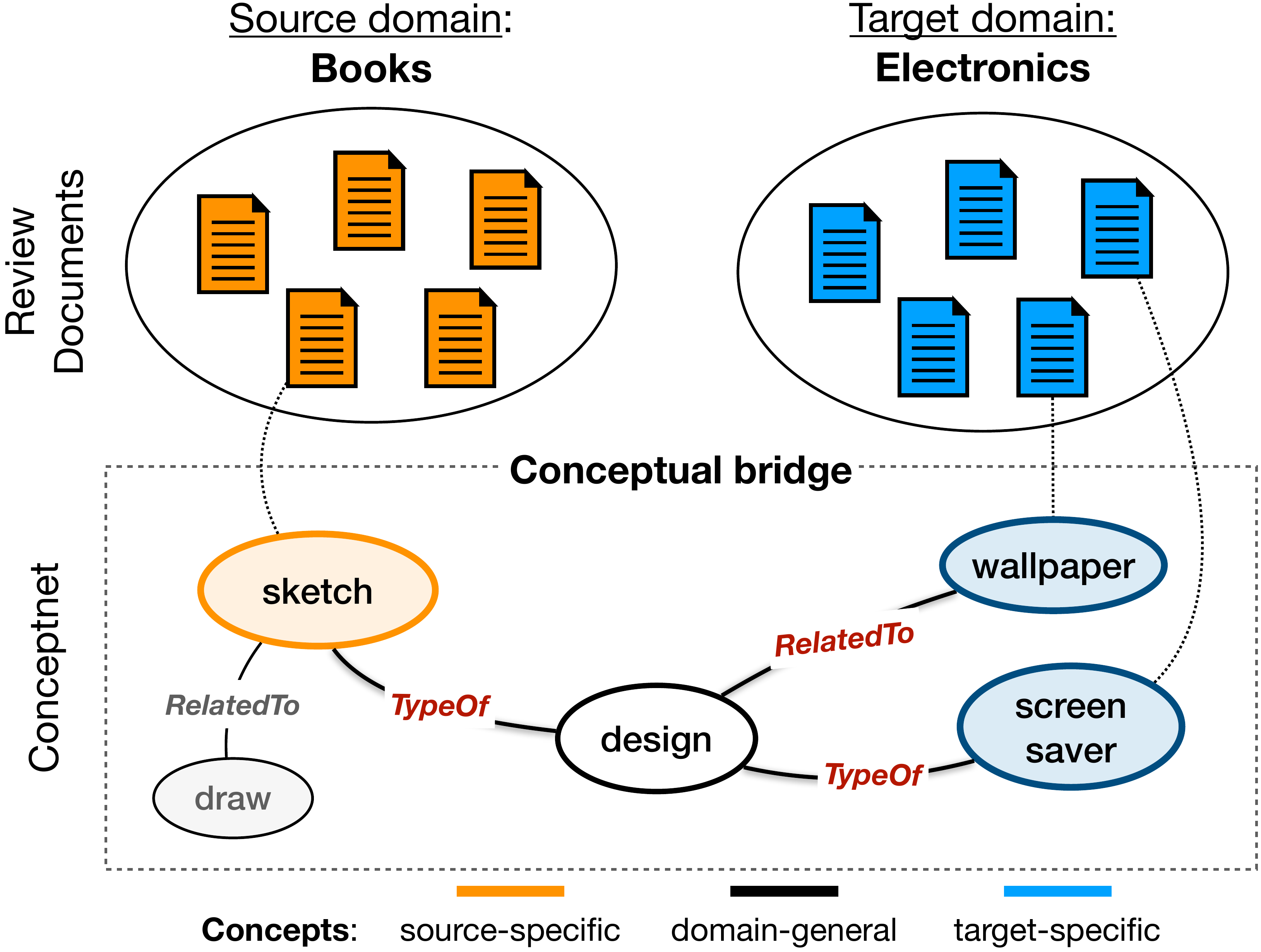}
    \caption{ \textit{ConceptNet} provides networks with background concepts that enhance their semantic understanding. For example, for a target sentence from electronics domain, \textit{The software came with decent \ul{screen savers}}, comprising domain-specific terms like \textit{screen saver} or \textit{wallpaper}, ConceptNet helps connecting them to general concepts like \textit{design}, thus allowing a network better understand their meaning. Furthermore, inter-domain conceptual bridge can also be established to connect source and target domains (\textit{wallpaper}--\textit{sketch} have similar conceptual notions under the link \textit{design}).}
    \label{fig:conceptual_links}
\end{figure}


Augmenting neural models with external knowledge bases (KB) has shown benefits across a range of NLP applications~\cite{peters-etal-2019-knowledge,li-etal-2019-improving,DBLP:conf/acl/LoganLPGS19,liu-etal-2019-knowledge-aware,bi-etal-2019-incorporating}. Despite their popularity, efforts to incorporate KBs into the domain-adaptation framework has been sporadic~\cite{DBLP:conf/icdm/WangDH08,DBLP:journals/tkde/XiangCHY10}. To this end, we identify multiple advantages of using commonsense KBs for domain adaptation. 

First, KBs help in grounding text to real entities, factual knowledge, and commonsense concepts. Commonsense KBs, in particular, provide a rich source of background concepts--related by commonsense links--which can enhance the semantics of a piece of text by providing both domain-specific and domain-general concepts~\cite{yang-etal-2019-enhancing-topic,zhong-etal-2019-knowledge,DBLP:journals/cin/AgarwalMBG15,zhong-etal-2019-knowledge} (see \cref{fig:conceptual_links}). For cross-domain SA, word polarities might vary among different domains. For example, {\it heavy} can be a positive feature for a truck, but a negative feature for a smartphone. It is, however, difficult to assign contextual-polarities solely from data, especially when there is no supervision~\cite{DBLP:conf/aaai/BoiaMF14}. In this domain-specific scenario, commonsense knowledge provides a dynamic way to enhance the context and help models understand sentiment-bearing terms and opinion targets through its structural relations~\cite{DBLP:conf/aaai/CambriaPHK18}. They also often aid in unearthing implicitly expressed sentiment~\cite{DBLP:conf/wassa/BalahurHM11}.


Second, domains often share relations through latent semantic concepts~\cite{DBLP:conf/icml/KimCKLK17}. For example, notions of \verb|wallpaper| (from electronics) and \verb|sketch| (from books) can be associated via related concepts such as \verb|design| (see~\cref{fig:conceptual_links}). Multi-relational KBs provide a natural way to leverage such inter-domain relationships. These connections can help models understand target-specific terms by associating to known domain-general or even source-specific concepts.



Following these intuitions, we propose a two-step modular framework, \textit{KinGDOM} (\textbf{K}nowledge-\textbf{G}uided \textbf{Dom}ain adaptation), which utilizes commonsense KB for domain adaptation. \textit{KinGDOM} first trains a shared graph autoencoder using a graph convolution network (GCN) on ConceptNet, so as to learn: 1) inter-domain conceptual links through multiple inference steps across neighboring concepts; and 2) domain-invariant concept representations due to shared autoencoding. It then extracts document-specific sub-graph embeddings and feeds them to a popular domain-adversarial model DANN~\cite{JMLR:v17:15-239}. Additionally, we also train a shared autoencoder on these extracted graph embeddings to promote further domain-invariance~\cite{DBLP:conf/icml/GlorotBB11}.

Our main contributions in this work are: 

\begin{enumerate}[leftmargin=*]
    \item We propose \textit{KinGDOM}, a domain-adversarial framework that uses an external KB (ConceptNet) for unsupervised domain adaptation. \textit{KinGDOM} learns domain-invariant features of KB concepts using a graph autoencoding strategy.
    \item We demonstrate, through experiments, that \textit{KinGDOM} surpasses state-of-the-art methods on the Amazon-reviews dataset~\cite{blitzer-etal-2007-biographies}, thus validating our claim that external knowledge can aid the task of cross-domain SA.
\end{enumerate}

In the remaining paper,~\cref{sec:related} explains related works and compares \textit{KinGDOM} to them;~\cref{sec:background} presents task definition and preliminaries;~\cref{sec:method} introduces our proposed framework, \textit{KinGDOM};~\cref{sec:exp} discusses experimental setup followed by results and extensive analyses in \cref{sec:results}; finally, \cref{sec:conclusion} concludes this paper.


\section{Related Work}
\label{sec:related}

Domain adaptation methods can be broadly categorized into three approaches: a) \textit{instance-selection}~\cite{DBLP:conf/acl/JiangZ07,DBLP:conf/nips/ChenWB11,DBLP:conf/cvpr/CaoL0J18}, b) \textit{self-labeling}~\cite{DBLP:journals/ipm/HeZ11} and c) \textit{representation learning}~\cite{DBLP:conf/icml/GlorotBB11,DBLP:conf/icml/ChenXWS12,DBLP:journals/corr/TzengHZSD14}. Our focus is on the third category which has emerged as a popular approach in this deep representation learning era~\cite{ruder2019neural,poria2020beneath}. 

\paragraph{Domain-adversarial Training.} Our work deals with domain-adversarial approaches~\cite{kouw2019review}, where we extend DANN~\citet{JMLR:v17:15-239}. Despite its popularity, DANN cannot model domain-specific information (e.g. indicators of \textit{tasty}, \textit{delicious} for kitchen domain)~\cite{DBLP:conf/acl/ZhangHPJ18}. Rectifications include shared-private encoders that model both domain-invariant and -specific features~\cite{DBLP:conf/aaai/LiJL12,DBLP:conf/nips/BousmalisTSKE16,DBLP:conf/acl/KimSK17a,DBLP:conf/cvpr/ChangWPC19}, using adversarial and orthogonality losses~\cite{DBLP:conf/acl/LiuQH17,DBLP:conf/naacl/LiBC18}. Although we do not use private encoders, we posit that our model is capable of capturing domain-specificity via the sentence-specific concept graph. Also, our approach is flexible enough to be adapted to the setup of shared-private encoders.



\paragraph{External Knowledge.}

Use of external knowledge has been explored in both inductive and transductive settings~\cite{DBLP:conf/icmla/Banerjee07,DBLP:conf/coling/DengSYLDFL18}. Few works have explored external knowledge in domain adaptation based on Wikipedia as auxiliary information, using co-clustering~\cite{DBLP:conf/icdm/WangDH08} and semi-supervised learning (SSL)~\cite{DBLP:journals/tkde/XiangCHY10}. SSL has also been explored by \citet{DBLP:conf/acl/JotyAI18} in the Twitter domain. Although we share a similar motivation, there exist crucial differences. Primarily, we learn graph embeddings at the concept level, not \textit{across} complete instances. Also, we do not classify each concept node in the graph, which renders SSL inapplicable to our setup.

\paragraph{Domain Adaptation on Graphs.}
With the advent of graph neural networks, graph-based methods have become a new trend~\cite{ghosal2019dialoguegcn} in diverse NLP tasks such as emotion recognition in conversations~\cite{poria2019emotion}.
Graph-based domain adaptation is categorized based on the availability of cross-domain connections. For domain-exclusive graphs, approaches include SSL with GCNs~\cite{DBLP:journals/corr/abs-1901-07264} and domain-adversarial learning~\cite{DBLP:journals/corr/abs-1909-01541}. For cross-domain connected graphs, co-regularized training~\cite{DBLP:conf/www/NiCLCCX018} and joint-embedding~\cite{DBLP:conf/wsdm/XuWCY17} have been explored. We also utilize GCNs to learn node representations in our cross-domain ConceptNet graph. However, rather than using explicit divergence measures or domain-adversarial losses for domain invariance, we uniquely adopt a shared-autoencoder strategy on GCNs. Such ideas have been explored in vector-based approaches~\cite{DBLP:conf/icml/GlorotBB11,DBLP:conf/icml/ChenXWS12}.

\paragraph{Sentiment Analysis.} 

One line of work models domain-dependent word embeddings~\cite{DBLP:conf/acl/SarmaLS18,DBLP:conf/acl/LamSBF18,k-sarma-etal-2019-shallow} or domain-specific sentiment lexicons~\cite{DBLP:conf/emnlp/HamiltonCLJ16}, while others attempt to learn representations based on co-occurrences of domain-specific with domain-independent terms~\cite{DBLP:conf/acl/BlitzerDP07, DBLP:conf/www/PanNSYC10, DBLP:conf/acl/BhattacharyyaDS18}. Our work is related to approaches that 
address domain-specificity in the target domain~\cite{DBLP:conf/acl/ZhangHPJ18,DBLP:conf/conll/BhattSR15}. Works like \citet{DBLP:conf/naacl/LiuZL18} attempts to model target-specificity by mapping domain-general information to domain-specific representations by using domain descriptor vectors.
In contrast, we address relating domain-specific terms by modeling their relations with the other terms in knowledge bases like ConceptNet.

\section{Background} 
\label{sec:background}

\subsection{Task Definition}
\label{subsec:task_def}

\begin{figure*}[t]
    \centering
    \includegraphics[width=\linewidth]{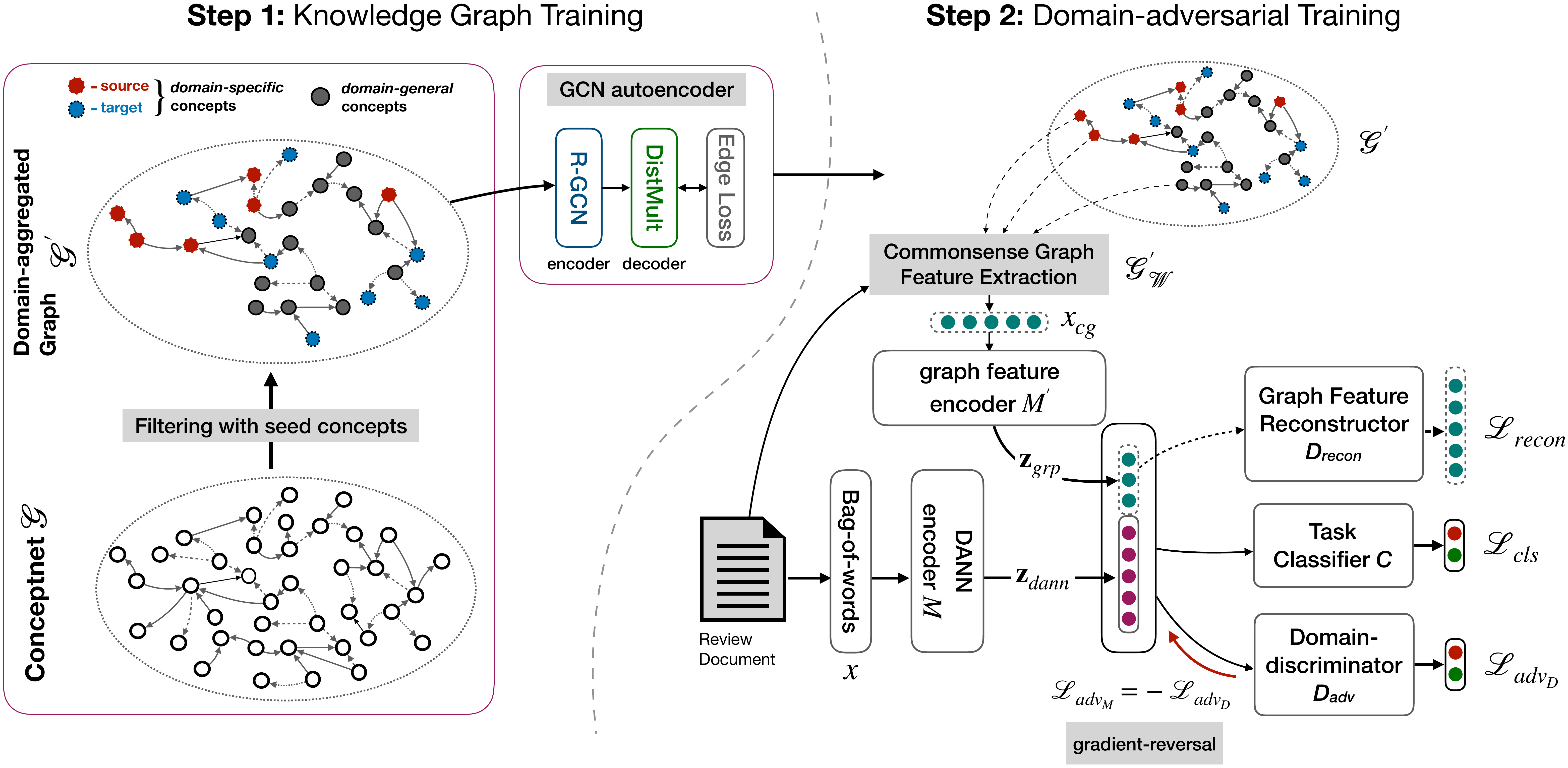}
    \caption{Illustration of \textit{KinGDOM}: Step 1 uses GCN to learn concept representations. Step 2 feeds concept features to DANN.}
    \label{fig:framework}
\end{figure*}

Domain adaptation deals with the training of models that can perform inference reliably in multiple domains. Across domains, it is assumed that the feature and label spaces are the same but with discrepancies in their feature distributions. In our setup, we consider two domains: source $\mathcal{D}_s$ and target domain $\mathcal{D}_t$ with different marginal data distributions, i.e., $P_{\mathcal{D}_s}(x) \neq P_{\mathcal{D}_t}(x)$. This scenario, also known as the \textit{covariate shift}~\cite{elsahar-galle-2019-annotate}, is predominant in SA applications and arises primarily with shifts in topics -- causing a difference in vocabulary usage and their corresponding semantic and sentiment associations.

We account for \textit{unsupervised} domain adaptation, where we are provided with labeled instances from the source domain $D_{s}^{l} = \{(x_i, y_i)\}_{i=1}^{N_s}$ and unlabeled instances from the target domain $D_{t}^{u} = \{(x_i)\}_{i=1}^{N_t}$.\footnote{For our case, each instance is a review document} This is a realistic setting as curating annotations for the target domain is often expensive as well as time consuming. Given this setup, our goal is to train a classifier that can achieve good classification performance on the target domain.

\subsection{Domain-Adversarial Neural Network}
\label{subsec:DANN}

We base our framework on the domain-adversarial neural network (DANN) proposed by~\citet{JMLR:v17:15-239}. DANN learns a shared mapping of both source and target domain instances $M(\mathbf{x}_{s/t})$ such that a classifier $C$ trained for the source domain can be directly applied for the target domain. Training of $C$ is performed using the cross-entropy loss:
\begin{align*}
    \begin{split}
        \mathcal{L}_{\mathrm{cls}} = 
        \mathbb{E}_{\left(x_{s}, y_{s}\right)} \left( -\sum_{k=1}^{K} \mathbb{1}_{\left[k=y_{s}\right]} \log C\left(M\left(x_{s}\right)\right)   \right), \label{eq:classifier_loss_dann}
    \end{split}
\end{align*}

\noindent where $K$ is the number of labels. Both the mapping function $M$ and the classifier $C$ are realized using neural layers with parameters $\theta_M$ and $\theta_C$.

\begin{table*}[ht!]
\scalebox{0.75}{
  \resizebox{\linewidth}{!}{
\begin{tabular}{lllll}
\textbf{All domains} & \multicolumn{1}{l}{\textbf{DVD}} & \multicolumn{1}{l}{\textbf{Books}} & \multicolumn{1}{l}{\textbf{Kitchen}} & \multicolumn{1}{l}{\textbf{Electronics}} \\ \hline 
RelatedTo (580k) & RelatedTo & RelatedTo & RelatedTo & RelatedTo \\
HasContext (80k) & HasContext & HasContext & IsA & IsA \\
IsA (60k) & IsA & IsA & Synonym & Synonym \\
DerivedFrom (42k) & Synonym & Synonym & DerivedFrom & DerivedFrom \\
Synonym (40k) & DerivedFrom & DerivedFrom & HasContext & HasContext \\
AtLocation (14k) & AtLocation & CapableOf & AtLocation & AtLocation \\
UsedFor (12k) & CapableOf & AtLocation & UsedFor & UsedFor \\
CapableOf (11k) & UsedFor & SimilarTo & SimilarTo & SimilarTo \\
SimilarTo (10k) & SimilarTo & UsedFor & CapableOf & CapableOf \\
Etymologically (5k) & Antonym & Antonym & Antonym & Antonym \\
\hline
\end{tabular}
}}
\caption{Top-10 relations of $\mathcal{G}^{'}$ based on frequency. Top relations for each domain are also mentioned.}
\label{tab:top_relations}
\end{table*}

\paragraph{Adversarial Loss.} The core idea of DANN is to reduce domain gap by learning common representations that are indistinguishable to a domain discriminator. To learn a domain-invariant mapping, DANN uses an adversarial discriminator $D_{adv}$ with parameters $\theta_D$, whose job is to distinguish between source and target instances, $M(x_s)$ vs. $M(x_t)$. It is trained using the cross-entropy loss:
\begin{align*}
    \begin{split}
        \mathcal{L}_{\mathrm{adv}_{D}} = 
        -\mathbb{E}_{\mathbf{x}_{s}} \left(\log D_{adv}\left(M\left(\mathbf{x}_{s}\right)\right)\right)  \\
        -\mathbb{E}_{\mathbf{x}_{t}}\left(\log \left(1-D_{adv}\left(M\left(\mathbf{x}_{t}\right)\right)\right)\right).
    \end{split}
\end{align*}

The mapping function then learns domain invariance by pitting against the discriminator in a minimax optimization with loss $\mathcal{L}_{\mathrm{adv}_{M}} = - \mathcal{L}_{\mathrm{adv}_{D}}$~\cite{DBLP:conf/cvpr/TzengHSD17}. This setup forces the features to become discriminative to the main learning task and indistinguishable across domains. The point estimates of the parameters are decided at a saddle point using the minimax objective:
\begin{align*}
    \begin{split}
        \theta^* = \argmin_{\theta_{M,C}} \max_{\theta_D} \left( \mathcal{L}_{\mathrm{cls}} + \lambda \, \mathcal{L}_{\mathrm{adv}_{D}} \right),
    \end{split}
\end{align*}
\noindent where $\lambda$ is a hyper-parameter. The minimax objective is realized by reversing the gradients of $\mathcal{L}_{adv_D}$ when back-propagating through $M$.

\section{Our Proposed Method} \label{sec:method}



\textit{KinGDOM} aims to improve the DANN approach by leveraging an external knowledge source i.e., ConceptNet. Such a knowledge base is particularly useful for domain adaptation as it contains both domain specific and domain general knowledge. Unlike traditional word embeddings and semantic knowledge graphs (e.g. WordNet), ConceptNet is unique as it contains commonsense related information. We posit that both these properties of  ConceptNet will be highly useful for domain adaptation. \textit{KinGDOM} follows a two-step approach described below:
\paragraph{Step 1:} This step deals with training a domain-aggregated sub-graph of ConceptNet. In particular, it involves: a) Creating a sub-graph of ConceptNet based on all domains (\cref{sec:subgraph_construction}). b) Training a graph-convolutional autoencoder to learn concept embeddings~\cite{DBLP:conf/esws/SchlichtkrullKB18} (\cref{sec:kb_training}).
\paragraph{Step 2:} After the graph autoencoder is trained, a) we extract and pool document-relevant features from the trained graph for each instance in the dataset (\cref{sec:kb_feature_extract}). b) The corresponding graph feature vector is then fed into the DANN architecture for adversarial training~\cite{JMLR:v17:15-239}. To further enforce domain invariance, we also introduce a shared autoencoder to reconstruct the graph features (\cref{sec:modified_DANN}).

\subsection{Step 1a) Domain-Aggregated Commonsense Graph Construction}
\label{sec:subgraph_construction}

We construct our domain-aggregated graph from ConceptNet~\cite{speer2017conceptnet}. First, we introduce the following notation: the ConceptNet graph is represented as a directed labeled graph $\mathcal{G = (V, E, R)}$, with concepts/nodes~\footnote{We use \textit{node}, \textit{concept}, and \textit{entity} interchangeably} $v_{i} \in \mathcal{V}$ and labeled edges $(v_i, r_{ij}, v_j)\in \mathcal{E}$, where $r_{ij} \in \mathcal{R}$ is the relation type of the edge between $v_i$ and $v_j$. 
The concepts in ConceptNet are unigram words or n-gram phrases. For instance one such triplet from ConceptNet is [\textit{baking-oven}, \textit{AtLocation}, \textit{kitchen}].

ConceptNet has approximately 34 million edges, from which we first extract a subset of edges. From the training documents of all domains in our dataset, we first extract the set of all the unique nouns, adjectives, and adverbs.\footnote{We use the Spacy POS Tagger: \url{https://spacy.io/usage/linguistic-features\#pos-tagging}} These extracted words are treated as the seeds that we use to filter ConceptNet into a sub-graph. In particular, we extract all the triplets from $\mathcal{G}$ which are within a distance of 1 to any of those seed concepts, resulting in a sub-graph $\mathcal{G' = (V', E', R')}$, with approximately $356k$ nodes and $900k$ edges. This sub-graph would thus contain concepts across all domains along with inter-concept links. Looking at the sub-graph $\mathcal{G'}$ from the lens of each domain, we can observe the top-10 relations within the domain in \cref{tab:top_relations}.



\subsection{Step 1b) Knowledge Graph Pre-training}
\label{sec:kb_training}

To utilize $\mathcal{G'}$ in our task, we first need to compute a representation of its nodes. We do this by training a graph autoencoder model to perform link prediction. The model takes as input an incomplete set of edges $\hat{\mathcal{E'}}$ from $\mathcal{E'}$ in $\mathcal{G'}$ and then assign scores to possible edges $(c_1, r, c_2)$, determining how likely are these edges to be in $\mathcal{E'}$.
Following \citet{DBLP:conf/esws/SchlichtkrullKB18}, our graph autoencoder model consists of: a R-GCN entity encoder and a DistMult scoring decoder.

\paragraph{Encoder Module.}

We employ the Relational Graph Convolutional Network (R-GCN) encoder from \citet{DBLP:conf/esws/SchlichtkrullKB18} as our graph encoder network. The power of this model comes from its ability to accumulate relational evidence in multiple inference steps from the local neighborhood around a given concept. The neighborhood-based convolutional feature transformation process always ensures that distinct domains are connected via underlying concepts and influence each other to create enriched domain-aggregated feature vectors.

Precisely, our encoder module consists of two R-GCN encoders stacked upon one another. The initial concept feature vector $\mathbf{g}_i$ is initialized randomly and thereafter transformed into the domain-aggregated feature vector $\mathbf{h}_{i} \in \mathbb{R}^d$ using the two-step graph convolution process. The transformation process is detailed below:
\begin{align*}
    f(\mathbf{x}_i , l)   = \sigma (\sum\limits_{r \in \mathcal{R}}^{} \sum\limits_{j \in N_{i}^{r}}^{} \frac{1}{c_{i,r}}W_{r}^{(l)} \mathbf{x}_{j} + W_{0}^{(l)}\mathbf{x}_{i}), \\
    \mathbf{h}_{i} = \mathbf{h}_{i}^{(2)}  = f(\mathbf{h}_i^{(1)},2) \quad ; \quad \mathbf{h}_{i}^{(1)} = f(\mathbf{g}_i \, 1),
\end{align*}

\noindent where $N_{i}^{r}$ denotes the neighbouring concepts of concept $i$ under relation $r \in \mathcal{R}$; $c_{i,r}$ is a normalization constant which either can be set in advance, such that, $c_{i,r} = |N_{i}^{r}|$, or can be learned in a gradient-based learning setup. $\sigma$ is an activation function such as ReLU, and $W_{r}^{(1/2)}$, $W_{0}^{(1/2)}$ are learnable parameters of the transformation.

This stack of transformations effectively accumulates the normalized sum of the local neighborhood i.e. the neighborhood information for each concept in the graph. The self-connection ensures self-dependent feature transformation.

\paragraph{Decoder Module.}
DistMult factorization \cite{yang2014embedding} is used as the scoring function. For a triplet $(c_i, r, c_j)$, the score $s$ is obtained as follows:
\begin{align*}
    \begin{split}
        s(c_i, r, c_j) = \sigma(\mathbf{h}_{c_i}^{T}R_{r}\mathbf{h}_{c_j}),
    \end{split}
\end{align*}
\noindent where $\sigma$ is the logistic function; $\mathbf{h}_{c_i}$, $\mathbf{h}_{c_j} \in \mathbb{R}^d$ are the R-GCN encoded feature vectors for concepts $c_i$, $c_j$. Each relation $r \in \mathcal{R}$ is also associated with a diagonal matrix $R_r \in \mathbb{R}^{d \times d}$.

\paragraph{Training.}
We train our graph autoencoder model using negative sampling~\cite{DBLP:conf/esws/SchlichtkrullKB18}. For triplets in $\hat{\mathcal{E'}}$ (positive samples), we create an equal number of negative samples by randomly corrupting the positive triplets. The corruption is performed by randomly modifying either one of the constituting concepts or the relation, creating the overall set of samples denoted by $\mathcal{T}$.

The task is set as a binary classification between the positive/negative triplets, where the model is trained with the standard cross-entropy loss:
\begin{align*}
    \begin{split}
        \mathcal{L}_{\mathcal{G'}} =-\frac{1}{2|\hat{\mathcal{E'}}|} \sum_{(c_i, r, c_j, y)\in \mathcal{T}}
        ( y\log s(c_i, r, c_j) + \\
        (1 - y)\log(1 - s(c_i, r, c_j))).
    \end{split}
\end{align*}

Once we train the autoencoder graph model, it will ensure that target domain-specific concepts (crucial for KG) can possibly be explained via domain-general concepts and further via inter-domain knowledge. In other words, the encoded node representations $\mathbf{h}_{i}$ will capture commonsense graph information in the form of domain-specific and domain-general features and thus will be effective for the downstream task when there is a distributional shift during evaluation.

\subsection{Step 2a) Commonsense Graph Feature Extraction}
\label{sec:kb_feature_extract}

The trained graph autoencoder model as explained in the previous section \cref{sec:kb_training}, can be used for  feature extraction. We now describe the methodology to extract the document-specific commonsense graph features for a particular document $x$: 
\begin{enumerate}[1),leftmargin=*]
    \item The first step is to extract the set of all unique nouns, adjectives, and adverbs present in the document. We call this set $\mathcal{W}$.
    \item Next, we extract a subgraph from $\mathcal{G'}$, where we take all triplets for which both the constituting nodes are either in $\mathcal{W}$ or are within the vicinity of radius 1 of any of the words in $\mathcal{W}$. We call this graph $\mathcal{G'_W}$. 
    \item We then make a forward pass of $\mathcal{G'_W}$ through the encoder of the pre-trained graph autoencoder model. This results in feature vectors $\mathbf{h}_j$ for all unique nodes $j$ in $\mathcal{G'_W}$. 
    \item Finally, we average over the feature vectors $\mathbf{h}_j$ for all unique nodes in $\mathcal{G'_W}$, to obtain the commonsense graph features $\mathbf{x}_{cg}$ for document $x$.
\end{enumerate}

We surmise that since most documents will have both domain-specific and domain-general words in $\mathcal{W}$, $\mathbf{x}_{cg}$ will inherently capture the commonsense information likely to be helpful during domain adaptation.

\subsection{Step 2b) Domain-adversarial Training}
\label{sec:modified_DANN}

We feed the commonsense graph feature $\mathbf{x}_{cg}$ pooled from $\mathcal{G'_W}$ for document $x$ (\cref{sec:kb_feature_extract}) into the DANN architecture (see ~\cref{subsec:DANN}). We proceed by learning a encoder function for the graph vector $\mathbf{z}_{grp} = M_{\theta_G}^{'}(\mathbf{x}_{cg})$ and combine its representation with the DANN encoder $\mathbf{z}_{dann} = M_{\theta_M}(\mathbf{x})$ to get the final feature representation $[\mathbf{z}_{dann};\mathbf{z}_{grp}]$, of the document $x$. Here, $[a;b]$ represents concatenation.

The task classifier $C$ and domain-discriminator $D_{adv}$ now takes this modified representation, $[\mathbf{z}_{dann};\mathbf{z}_{grp}]$, as its input instead of only $\mathbf{z}_{dann}$. To further enforce domain-invariance into the encoded graph representation $\mathbf{z}_{grp}$, we consider it as a hidden code in a traditional autoencoder and consequently add a shared decoder $D_{recon}$ (with parameters $\theta_{R}$) with a reconstruction loss (mean-squared error):
\begin{align*}
    \begin{split}
        \mathcal{L}_{\mathrm{recon}}\left(X_{s}, X_{t} \right) =  \mathcal{L}_{\mathrm{recon}}\left(X_{s} \right) + \mathcal{L}_{\mathrm{recon}}\left(X_{t} \right),\\
        \text{s.t.} \quad \mathcal{L}_{\mathrm{recon}} = -\mathbb{E}_{\mathbf{x_{cg}}} \left( \norm{ D_{recon}(\mathbf{z}_{grp}) - \mathbf{x_{cg}} }_2^2 \right).
    \end{split}
\end{align*}

We hypothesize that if $\theta_{R}$ can reconstruct graph features for both domains, then it would ensure stronger domain invariance constraints in $\mathbf{z}_{grp}$.
The final optimization of this domain-adversarial setup is based on the minimax objective:
\begin{align*}
    \begin{split}
        \theta^* = \argmin_{\theta_{G, M, C, R}} \max_{\theta_D} \left( \mathcal{L}_{\mathrm{cls}} + \lambda \, \mathcal{L}_{\mathrm{adv}_{D}} + \gamma  \, \mathcal{L}_{\mathrm{recon}} \right),
    \end{split}
\end{align*}
where $\lambda$ and $\gamma$ are hyper-parameters.

\begin{figure*}[t]
    \centering
    \includegraphics[width=0.98\linewidth]{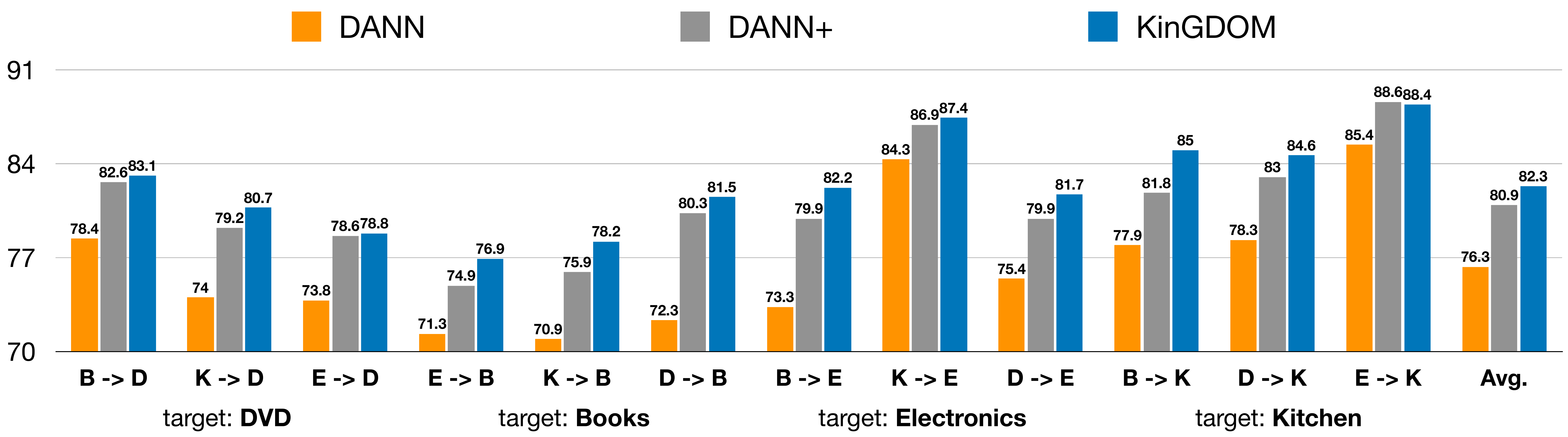}
    \caption{Results of DANN vs DANN+ vs \textit{KinGDOM} across different target domains.
    Best viewed in colour.}
    \label{fig:dann_compare}
\end{figure*}

\section{Experimental Setup} \label{sec:exp}

\subsection{Dataset} \label{subsec:dataset}

We consider the Amazon-reviews benchmark dataset for domain adaptation in SA~\cite{blitzer-etal-2007-biographies}. This corpus consists of Amazon product reviews and ranges across four domains: Books, DVDs, Electronics, and Kitchen appliances. Each review is associated with a rating denoting its sentiment polarity. Reviews with rating up to $3$ stars are considered to contain negative sentiment and $4$ or $5$ stars as positive sentiment. The dataset follows a balanced distribution between both labels yielding $2k$ unlabelled training instances for each domain. Testing contains $3k$ - $6k$ samples for evaluation. We follow similar pre-processing as bone by ~\citet{JMLR:v17:15-239,DBLP:conf/acl/PlankR18} where each review is encoded into a $5000$-dimensional tf-idf weighted bag-of-words (BOW) feature vector of unigrams and bigrams.

\subsection{Training Details} \label{sec:training}

We follow~\citet{JMLR:v17:15-239} in training our network.
Our neural layers i.e., DANN encoder ($M$), graph feature encoder ($M'$), graph feature reconstructor ($D_{recon}$), task classifier ($C$) and domain discriminator ($D_{adv}$) are implemented with 100 dimensional fully connected layers. We use a cyclic $\lambda$ as per~\cite{JMLR:v17:15-239} and $\gamma = 1$ after validating with $\gamma \in \{0.5, 1, 2\}$. 25\% dropout is used in the fully connected layers and the model is trained with Adam~\cite{DBLP:journals/corr/KingmaB14} optimizer.

\subsection{Baseline Methods} 
\label{sec:baseline}

In this paper, to inspect the role of external commonsense knowledge and analyze the improvement in performance it brings, we intentionally use BOW features and compare them against other baseline models that also use BOW features. This issue has also been addressed by \citet{poria2020beneath}. The flexibility of \textit{KinGDOM} allows other approaches, such as mSDA, CNN, etc. to be easily incorporated in it, which we plan to analyze in the future.

We compare \textit{KinGDOM} with the following unsupervised domain adaptation baseline methods: \textbf{DANN}~\cite{JMLR:v17:15-239} is a domain-adversarial method, based on which we develop \textit{KinGDOM} (\cref{subsec:DANN}); \textbf{DANN+} The DANN model where we use an Adam optimizer instead of the original SGD optimizer. The network architecture and the rest of the hyperparameters are kept same; Variational Fair Autoencoder  (\textbf{VFAE})~\cite{louizos2015variational} learns latent representations independent from sensitive domain knowledge, while  retaining enough task information by using a MMD-based loss; Central Moment Discrepancy (\textbf{CMD})~\cite{zellinger2017central} is a regularization method which minimizes the difference between feature representations by utilizing equivalent representation of probability distributions by moment sequences; \textbf{Asym}~\cite{saito2017asymmetric} is the asymmetric tri-training framework that uses three neural networks asymmetrically for domain adaptation; \textbf{MT-Tri}~\cite{DBLP:conf/acl/PlankR18} is similar to \textbf{Asym}, but uses multi-task learning; Domain Separation Networks (\textbf{DSN})~\cite{bousmalis2016domain} learns to extract shared and private components of each domain. As per \citet{peng2018cross}, it stands as the present state-of-the-art method for unsupervised domain adaptation; Task Refinement Learning (\textbf{TRL})~\cite{ziser2019task} Task Refinement Learning is an unsupervised domain adaptation framework which iteratively trains a Pivot Based Language Model to gradually increase the information exposed about each pivot; \textbf{TAT}~\cite{liu2019transferable} is the transferable adversarial training setup to generate examples which helps in modelling the domain shift. 
TAT adversarially trains classifiers to make consistent predictions over these transferable examples; \textbf{CoCMD}~\cite{peng2018cross} is a co-training method based on the CMD regularizer which trains a classifier on simultaneously extracted domain specific and invariant features. CoCOMD, however, is SSL-based as it uses labeled data from the target domain. Although it falls outside the regime of unsupervised domain adaptation, we report its results to provide a full picture to the reader.

\begin{table*}[t]
\scalebox{1.0}{
  \resizebox{\linewidth}{!}{
  \centering
  \begin{tabular}{l||l|l|l|l|l|l|l|l|l|l||l|l|l}
    Source & & & & & & & & & & & & &\\
    $\quad \downarrow$ & & & & & & & & & & & & &\\
    Target & \rotatebox{90}{\parbox{2mm}{\multirow{3}{*}{DANN (5k)}}} &
    \rotatebox{90}{\parbox{2mm}{\multirow{3}{*}{DANN+ (5k)}}} &
    \rotatebox{90}{\parbox{2mm}{\multirow{3}{*}{VFAE (5k)}}} & \rotatebox{90}{\parbox{2mm}{\multirow{3}{*}{CMD (5k)}}} & \rotatebox{90}{\parbox{2mm}{\multirow{3}{*}{Asym (5k)}}} & \rotatebox{90}{\parbox{2mm}{\multirow{3}{*}{MT-Tri (5k)}}} & \rotatebox{90}{\parbox{2mm}{\multirow{3}{*}{TRL* (5k)}}} &
    \rotatebox{90}{\parbox{2mm}{\multirow{3}{*}{DSN (5k)}}} &
    \rotatebox{90}{\parbox{2mm}{\multirow{3}{*}{CoCMD* (5k)}}} &
    \rotatebox{90}{\parbox{2mm}{\multirow{3}{*}{\textbf{\footnotesize{KinGDOM (5k)}}}}}&
    \rotatebox{90}{\parbox{2mm}{\multirow{3}{*}{DANN+ (30k)}}}&
    \rotatebox{90}{\parbox{2mm}{\multirow{3}{*}{TAT (30k)}}} &
    \rotatebox{90}{\parbox{2mm}{\multirow{3}{*}{\textbf{\footnotesize{KinGDOM (30k)}}}}}
    \\
    \hline
    \hline
    B $\rightarrow$ D & 78.4 & 82.6 & 79.9 & 80.5 & 80.7 & 81.2 & 82.2 & 82.8 & 83.1 & 83.1 & 84.7 & 84.5 & \bf{85.0} \\
    B $\rightarrow$ E & 73.3 & 79.9 & 79.2 & 78.7 & 79.8 & 78.0 & - & 81.9 & 83.0 & 82.2 & 83.0 & 80.1 & \bf{83.9} \\
    B $\rightarrow$ K & 77.9 & 81.8 & 81.6 & 81.3 & 82.5 & 78.8 & 82.7 & 84.4 &85.3 & 85.0 & 84.0 & 83.6 & \bf{86.6}  \\
    D $\rightarrow$ B & 72.3 & 80.3 & 75.5 & 79.5 & 73.2 & 77.1 & - & 80.1 & 81.8 & 81.4 & 82.7 & 81.9 & \bf{82.7} \\
    D $\rightarrow$ E & 75.4 & 79.9 & 78.6 & 79.7 & 77.0 & 81.0 & - & 81.4 & 83.4 & 81.7 & 83.4 & 81.9 & \bf{83.9} \\
    D $\rightarrow$ K & 78.3 & 83.0 & 82.2 & 83.0 & 82.5 & 79.5 & - & 83.3 & 85.5 & 84.6 & 85.3 & 84.0 & \bf{87.1}  \\
    E $\rightarrow$ B & 71.3 & 74.9 & 72.7 & 74.4 & 73.2 & 73.5 & - & 75.1 & 76.9 & 76.9 & 77.1 & \bf{83.2} & 78.4  \\
    E $\rightarrow$ D & 73.8 & 78.6 & 76.5 & 76.3 & 72.9 & 75.4 & 75.8 & 77.1 &78.3 & 78.8 & 79.6 & 77.9 & \bf{80.3} \\
    E $\rightarrow$ K & 85.4 & 88.6 & 85.0 & 86.0 & 86.9 & 87.2 & - & 87.2 &87.3 & 88.4 & 89.0 & \bf{90.0} & 89.4 \\
    K $\rightarrow$ B & 70.9 & 75.9 & 72.0 & 75.6 & 72.5 &  73.8 & 72.1 & 76.4 &77.2 & 78.2 & 77.1 & 75.8 & \bf{80.0} \\
    K $\rightarrow$ D & 74.0 & 79.2 & 73.3 & 77.5 & 74.9 &  77.8 & - & 78.0 &79.6 & 80.7 & 81.3 & 77.7 & \bf{82.3} \\
    K $\rightarrow$ E & 84.3 & 86.9 & 83.8 & 85.4 & 84.6 &  86.0 & - & 86.7 & 87.2 & 87.4 & 88.0 & 88.2 & \bf{88.6} \\
    \hline
    \multicolumn{1}{c||}{Avg.}& 76.3 & 80.9 & 78.4 & 79.8 & 78.4 & 79.1 & - & 81.2 &82.4 & 82.3 & 82.9 & 82.4 & \textbf{84.0}\\
    \hline
  \end{tabular}
  }}
  \caption{Comparison with different baseline and state-of-the-art models (\cref{sec:baseline}). TRL* reported results on four combinations. CoCMD* is a semi-supervised domain adaptation method. DSN is the current state-of-the-art for unsupervised domain adaptation on the Amazon reviews dataset. Scores for MT-Tri are extrapolated from the graphs illustrated in~\citet{DBLP:conf/acl/PlankR18}. Note: \textbf{B}: Books, \textbf{D}: DVD, \textbf{E}:Electronics, and \textbf{K}: Kitchen domains. 5k, 30k signify 5000 and 30,000 dimensional BOW features. 
}
  \label{results:1}
\end{table*}


\section{Results and Analysis} \label{sec:results}

As mentioned in \cref{sec:baseline}, we reimplemented the baseline DANN model using Adam optimizer and observed that its results has been notably under-reported in many of the unsupervised domain adaptation literature for sentiment analysis (see \cref{results:1}). In the original DANN implementation~\cite{JMLR:v17:15-239}, Stochastic Gradient Descent (SGD) was used as the optimizer. However, in DANN+, using Adam optimizer leads to substantial jump in performance which can comfortably surpass many of the recent advanced domain adaptation methods -- CMD~\cite{zellinger2017central}, VFAE~\cite{louizos2015variational}, ASym~\cite{saito2017asymmetric} and MT-Tri~\cite{DBLP:conf/acl/PlankR18}.

We first compare the performance of \textit{KinGDOM} with its base models -- DANN and DANN+. As observed in~\cref{fig:dann_compare}, \textit{KinGDOM} surpasses DANN+ by 1.4\% which asserts the improvement in domain-invariance due to the incorporation of external commonsense knowledge.

Next, we look at ~\cref{results:1} where comparisons are made with other baselines, including the state-of-the-art DSN approach. As observed, \textit{KinGDOM} outperforms DSN in all the task scenarios, indicating the efficacy of our approach. \citet{blitzer-etal-2007-biographies}, in their original work, noted that domain transfer across the two groups of {\textit{DVD}, \textit{Books}} and {\textit{Electronics}, \textit{Kitchen}} is particularly challenging. Interestingly, in our results, we observe the highest gains when the source and target domains are from these separate groups (e.g., Kitchen $\rightarrow$ DVD, Kitchen $\rightarrow$ Books, Electronics $\rightarrow$ Books). 

In ~\cref{results:1}, we also compare \textit{KinGDOM} against CoCMD and TAT. Although CoCMD is a semi-supervised method, \textit{KinGDOM} surpasses its performance in several of the twelve domain-pair combinations and matches its overall result without using any labelled samples from the target domain. TAT is the state-of-the-art method for unsupervised domain adaptation in the Amazon reviews dataset when used with 30,000 Bag-Of-Words (BOW) features. Interestingly, \textit{KinGDOM} used with 5000 BOW features can match TAT with 30,000 BOW features and outperforms TAT by around 1.6\% overall when used with the same 30,000 BOW features. The reimplementation of DANN -- DANN+ with 30,000 BOW also surpasses the result of TAT by 0.5\%.
The results indicate that external knowledge, when added to a simple architecture such as DANN, can surpass sophisticated state-of-the-art models, such as DSN and TAT. Our primary intention to utilize DANN as the base model is to highlight the role of knowledge base infusion in domain adaptation, devoid of sophisticated models, and complex neural maneuvering. Nevertheless, the flexibility of \textit{KinGDOM} allows it to be associated with advanced models too (e.g., DSN, TAT), which we believe could perform even better. We intend to analyze this in the future.

\subsection{Ablation Studies} \label{sec:ablation}

We further analyze our framework and challenge our design choices. Specifically, we consider three variants of our architecture based on alternative ways to condition DANN with the graph features. Each of these variants reveals important clues regarding the invariance properties and task appropriateness of $\mathbf{z}_{grp}$. \textbf{Variant 1} denotes separate decoders $D_{recon}$ for source and target domains. In \textbf{Variant 2}, domain classifier $D_{adv}$ takes only $\mathbf{z}_{dann}$ as input whereas the sentiment classifier $C$ takes the concatenated feature $[\mathbf{z}_{dann};\mathbf{z}_{grp}]$. Finally, in \textbf{Variant 3}, $D_{adv}$ takes input $[\mathbf{z}_{dann};\mathbf{z}_{grp}]$ whereas $C$ only takes $\mathbf{z}_{dann}$.
\begin{figure}[ht!]
    \centering
    \includegraphics[width=\linewidth]{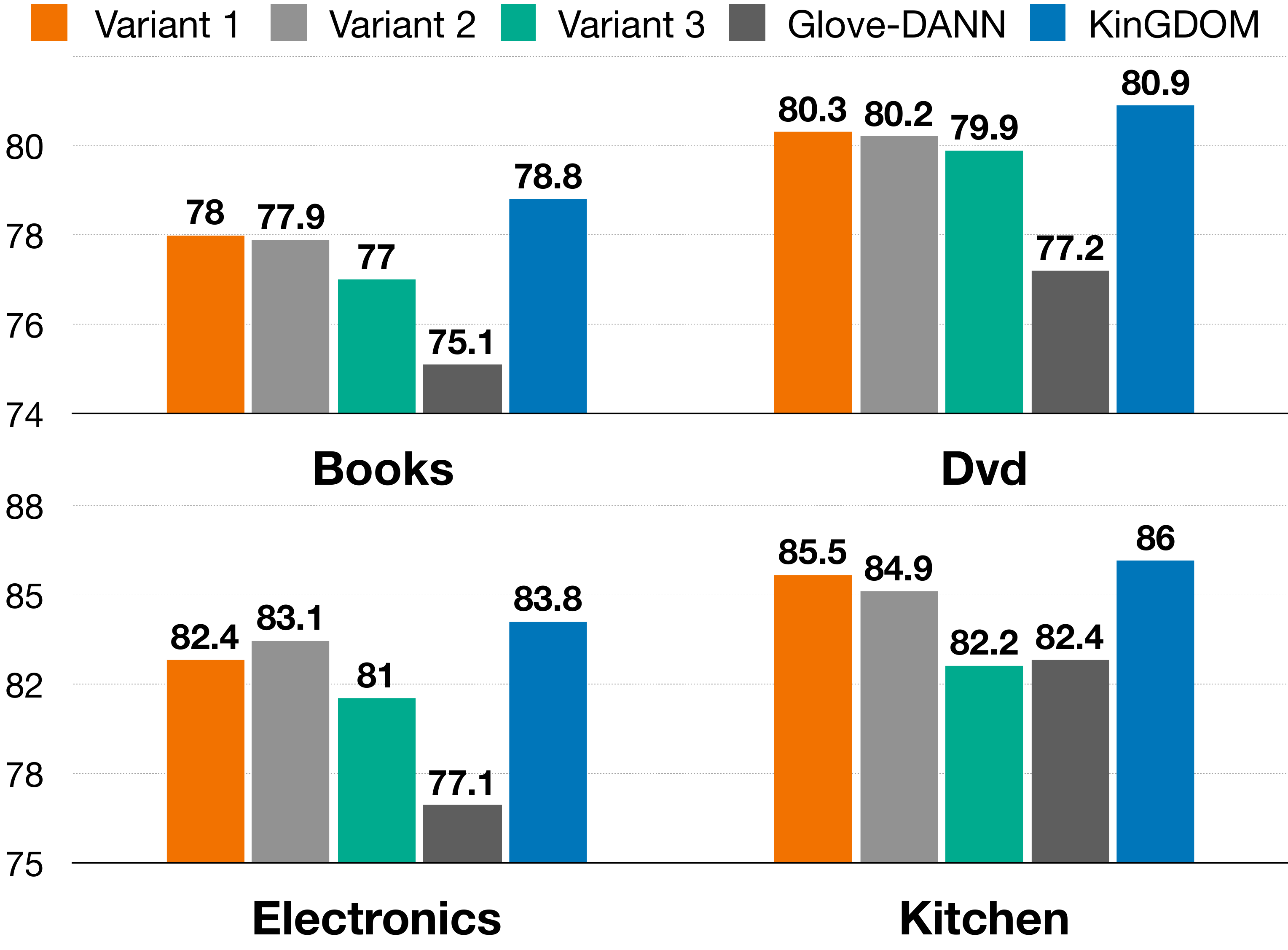}
    \caption{Average accuracy $(\%)$ on target domains across different variants defined in \cref{sec:ablation}. Best viewed in colour.}
    \label{fig:variants}
\end{figure}
As seen in \cref{fig:variants}, all the variants perform worse than \textit{KinGDOM}. For Variant 1, the performance drop indicates that having a shared decoder $D_{recon}$ in \textit{KinGDOM} facilitates learning invariant representations and helps target domain classification. For Variant 2, removal of $\mathbf{z}_{grp}$ from domain classifier diminishes the domain-invariance capabilities, thus making the domain classifier stronger and leading to a drop in sentiment classification performance. For Variant 3, removal of $\mathbf{z}_{grp}$ from sentiment classifier $C$ degrades the performance. This indicates that in \textit{KinGDOM}, $\mathbf{z}_{grp}$ contain task appropriate features retrieved from external knowledge (see \cref{section:intro}).

Besides ablations, we also look at alternatives to the knowledge graph and bag-of-words representation used for the documents. For the former, we consider replacing ConceptNet with WordNet~\cite{fellbaum2010wordnet}, which is a lexical knowledge graph with conceptual-semantic and lexical connections. We find the performance of \textit{KinGDOM} with WordNet to be $1\%$ worse than ConceptNet in terms of average accuracy score. This indicates the compatibility of ConceptNet with our framework. However, the competitive performance with WordNet also suggests the usability of our framework with any structural resource comprising inter-domain connections. For the latter, we use Glove-averaged embeddings with DANN. Glove is a popular word embedding method which captures semantics using co-occurrence statistics~\cite{DBLP:conf/emnlp/PenningtonSM14}. Results in \cref{fig:variants} show that using only Glove does not provide the amount of conceptual semantics available in ConceptNet.


\subsection{Case Studies}
\label{sec:case_study}

\begin{figure}[ht!]
    \centering
    \includegraphics[width=\linewidth]{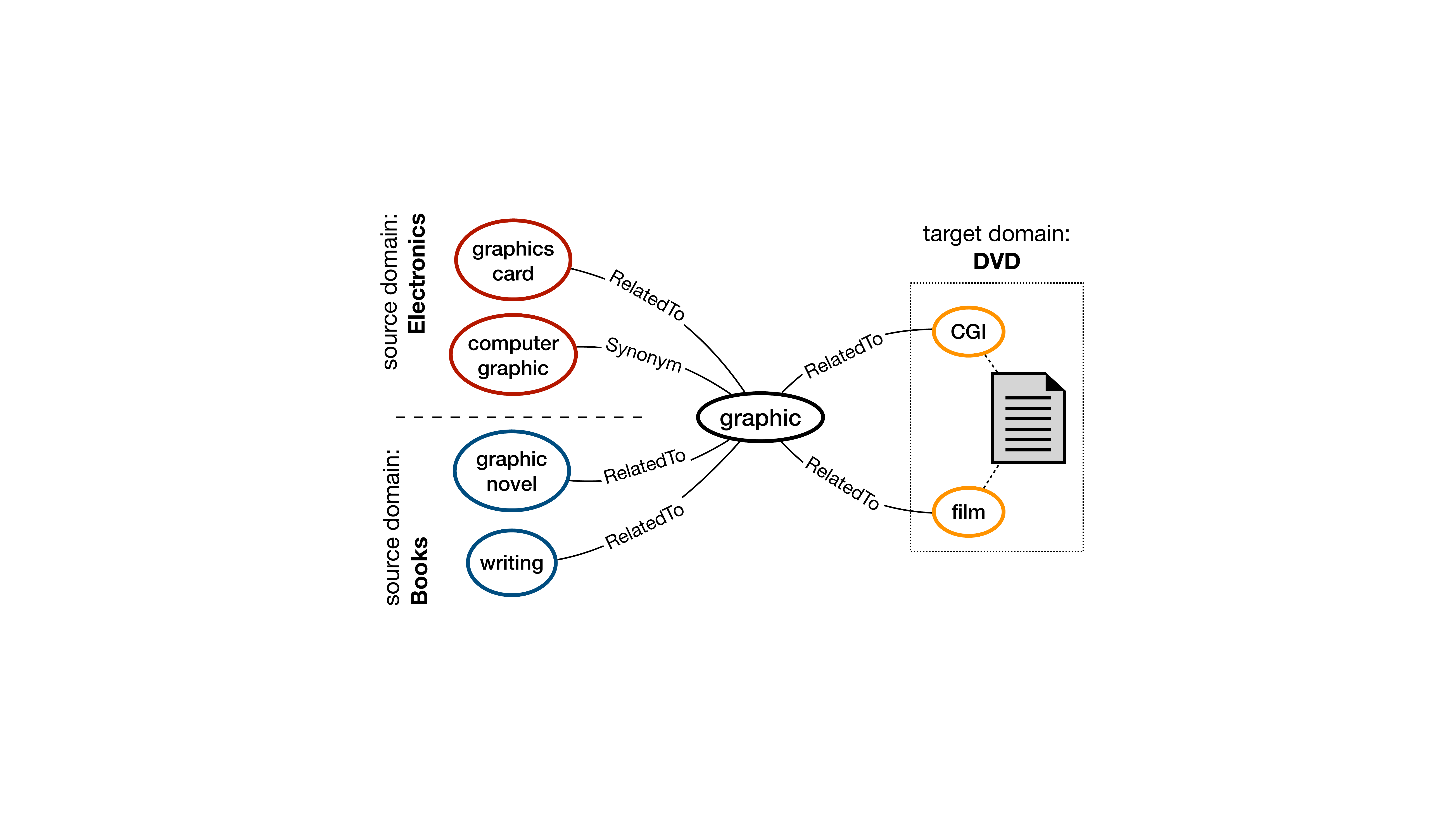}
    \caption{Domain-general term \textit{graphic} bridges the commonsense knowledge between domain-specific terms in Electronics, Books and DVD.}
    \label{fig:case_study}
\end{figure}

We delve further into our results and qualitatively analyze \textit{KinGDOM}. We look at a particular test document from DVD domain, for which \textit{KinGDOM} predicts the correct sentiment, both when the source domain is Electronics and also Books. In similar settings, DANN mispredicts the same document. Looking at the corresponding document-specific sub-graph for this document, we observe conceptual links to both domain-general concepts and domain-specific concepts from the source domain. In \cref{fig:case_study}, we can see the domain-specific terms \verb|CGI| and \verb|film| to be related to the general concept \verb|graphic| which is further linked to domain-specific concepts like \verb|graphics card|, \verb|writing|, etc. from Electronics, Books, respectively. This example shows how \textit{KinGDOM} might use these additional concepts to enhance the semantics as required for sentiment prediction.

\section{Conclusion} \label{sec:conclusion}

In this paper, we explored the role of external commonsense knowledge for domain adaptation. We introduced a domain-adversarial framework called {\it KinGDOM}, which relies on an external commonsense KB (ConceptNet) to perform unsupervised domain adaptation. We showed that we can learn domain-invariant features for the concepts in the KB by using a graph convolutional autoencoder. Using the standard Amazon benchmark for domain adaption in sentiment analysis, we showed that our framework exceeds the performance of previously proposed methods for the same task. Our experiments demonstrate the usefulness of external knowledge for the task of cross-domain sentiment analysis. 
Our code is publicly available at \url{https://github.com/declare-lab/kingdom}.

\section*{Acknowledgments}
This research is supported by A*STAR under its RIE 2020 Advanced Manufacturing and Engineering (AME) programmatic grant, Award No. -  A19E2b0098. 

\bibliography{acl2020}

\begin{thebibliography}{71}
\expandafter\ifx\csname natexlab\endcsname\relax\def\natexlab#1{#1}\fi

\bibitem[{Agarwal et~al.(2015)Agarwal, Mittal, Bansal, and
  Garg}]{DBLP:journals/cin/AgarwalMBG15}
Basant Agarwal, Namita Mittal, Pooja Bansal, and Sonal Garg. 2015.
\newblock \href {https://doi.org/10.1155/2015/715730} {Sentiment analysis using
  common-sense and context information}.
\newblock \emph{Comp. Int. and Neurosc.}, 2015:715730:1--715730:9.

\bibitem[{Alam et~al.(2018)Alam, Joty, and Imran}]{DBLP:conf/acl/JotyAI18}
Firoj Alam, Shafiq~R. Joty, and Muhammad Imran. 2018.
\newblock \href {https://doi.org/10.18653/v1/P18-1099} {Domain adaptation with
  adversarial training and graph embeddings}.
\newblock In \emph{Proceedings of the 56th Annual Meeting of the Association
  for Computational Linguistics, {ACL} 2018, Melbourne, Australia, July 15-20,
  2018, Volume 1: Long Papers}, pages 1077--1087.

\bibitem[{Balahur et~al.(2011)Balahur, Hermida, and
  Montoyo}]{DBLP:conf/wassa/BalahurHM11}
Alexandra Balahur, Jes{\'{u}}s~M. Hermida, and Andr{\'{e}}s Montoyo. 2011.
\newblock \href {https://www.aclweb.org/anthology/W11-1707/} {Detecting
  implicit expressions of sentiment in text based on commonsense knowledge}.
\newblock In \emph{Proceedings of the 2nd Workshop on Computational Approaches
  to Subjectivity and Sentiment Analysis, WASSA@ACL 2011, Portland, OR, USA,
  June 24, 2011}, pages 53--60. Association for Computational Linguistics.

\bibitem[{Banerjee(2007)}]{DBLP:conf/icmla/Banerjee07}
Somnath Banerjee. 2007.
\newblock \href {https://doi.org/10.1109/ICMLA.2007.39} {Boosting inductive
  transfer for text classification using wikipedia}.
\newblock In \emph{The Sixth International Conference on Machine Learning and
  Applications, {ICMLA} 2007, Cincinnati, Ohio, USA, 13-15 December 2007},
  pages 148--153. {IEEE} Computer Society.

\bibitem[{Bhatt et~al.(2015)Bhatt, Semwal, and Roy}]{DBLP:conf/conll/BhattSR15}
Himanshu~Sharad Bhatt, Deepali Semwal, and Shourya Roy. 2015.
\newblock \href {https://www.aclweb.org/anthology/K15-1006/} {An iterative
  similarity based adaptation technique for cross-domain text classification}.
\newblock In \emph{Proceedings of the 19th Conference on Computational Natural
  Language Learning, CoNLL 2015, Beijing, China, July 30-31, 2015}, pages
  52--61.

\bibitem[{Bi et~al.(2019)Bi, Wu, Yan, Wang, Xia, and
  Li}]{bi-etal-2019-incorporating}
Bin Bi, Chen Wu, Ming Yan, Wei Wang, Jiangnan Xia, and Chenliang Li. 2019.
\newblock \href {https://doi.org/10.18653/v1/D19-1255} {Incorporating external
  knowledge into machine reading for generative question answering}.
\newblock In \emph{Proceedings of the 2019 Conference on Empirical Methods in
  Natural Language Processing and the 9th International Joint Conference on
  Natural Language Processing (EMNLP-IJCNLP)}, pages 2521--2530, Hong Kong,
  China. Association for Computational Linguistics.

\bibitem[{Blitzer et~al.(2007{\natexlab{a}})Blitzer, Dredze, and
  Pereira}]{DBLP:conf/acl/BlitzerDP07}
John Blitzer, Mark Dredze, and Fernando Pereira. 2007{\natexlab{a}}.
\newblock \href {https://www.aclweb.org/anthology/P07-1056/} {Biographies,
  bollywood, boom-boxes and blenders: Domain adaptation for sentiment
  classification}.
\newblock In \emph{{ACL} 2007, Proceedings of the 45th Annual Meeting of the
  Association for Computational Linguistics, June 23-30, 2007, Prague, Czech
  Republic}.

\bibitem[{Blitzer et~al.(2007{\natexlab{b}})Blitzer, Dredze, and
  Pereira}]{blitzer-etal-2007-biographies}
John Blitzer, Mark Dredze, and Fernando Pereira. 2007{\natexlab{b}}.
\newblock \href {https://www.aclweb.org/anthology/P07-1056} {Biographies,
  {B}ollywood, boom-boxes and blenders: Domain adaptation for sentiment
  classification}.
\newblock In \emph{Proceedings of the 45th Annual Meeting of the Association of
  Computational Linguistics}, pages 440--447, Prague, Czech Republic.
  Association for Computational Linguistics.

\bibitem[{Boia et~al.(2014)Boia, Musat, and Faltings}]{DBLP:conf/aaai/BoiaMF14}
Marina Boia, Claudiu~Cristian Musat, and Boi Faltings. 2014.
\newblock \href {http://www.aaai.org/ocs/index.php/AAAI/AAAI14/paper/view/8244}
  {Acquiring commonsense knowledge for sentiment analysis through human
  computation}.
\newblock In \emph{Proceedings of the Twenty-Eighth {AAAI} Conference on
  Artificial Intelligence, July 27 -31, 2014, Qu{\'{e}}bec City, Qu{\'{e}}bec,
  Canada}, pages 901--907. {AAAI} Press.

\bibitem[{Bousmalis et~al.(2016{\natexlab{a}})Bousmalis, Trigeorgis, Silberman,
  Krishnan, and Erhan}]{DBLP:conf/nips/BousmalisTSKE16}
Konstantinos Bousmalis, George Trigeorgis, Nathan Silberman, Dilip Krishnan,
  and Dumitru Erhan. 2016{\natexlab{a}}.
\newblock \href {http://papers.nips.cc/paper/6254-domain-separation-networks}
  {Domain separation networks}.
\newblock In \emph{Advances in Neural Information Processing Systems 29: Annual
  Conference on Neural Information Processing Systems 2016, December 5-10,
  2016, Barcelona, Spain}, pages 343--351.

\bibitem[{Bousmalis et~al.(2016{\natexlab{b}})Bousmalis, Trigeorgis, Silberman,
  Krishnan, and Erhan}]{bousmalis2016domain}
Konstantinos Bousmalis, George Trigeorgis, Nathan Silberman, Dilip Krishnan,
  and Dumitru Erhan. 2016{\natexlab{b}}.
\newblock Domain separation networks.
\newblock In \emph{Advances in neural information processing systems}, pages
  343--351.

\bibitem[{Cambria et~al.(2018)Cambria, Poria, Hazarika, and
  Kwok}]{DBLP:conf/aaai/CambriaPHK18}
Erik Cambria, Soujanya Poria, Devamanyu Hazarika, and Kenneth Kwok. 2018.
\newblock \href
  {https://www.aaai.org/ocs/index.php/AAAI/AAAI18/paper/view/16839} {Senticnet
  5: Discovering conceptual primitives for sentiment analysis by means of
  context embeddings}.
\newblock In \emph{Proceedings of the Thirty-Second {AAAI} Conference on
  Artificial Intelligence, (AAAI-18), the 30th innovative Applications of
  Artificial Intelligence (IAAI-18), and the 8th {AAAI} Symposium on
  Educational Advances in Artificial Intelligence (EAAI-18), New Orleans,
  Louisiana, USA, February 2-7, 2018}, pages 1795--1802. {AAAI} Press.

\bibitem[{Cao et~al.(2018)Cao, Long, Wang, and
  Jordan}]{DBLP:conf/cvpr/CaoL0J18}
Zhangjie Cao, Mingsheng Long, Jianmin Wang, and Michael~I. Jordan. 2018.
\newblock \href {https://doi.org/10.1109/CVPR.2018.00288} {Partial transfer
  learning with selective adversarial networks}.
\newblock In \emph{2018 {IEEE} Conference on Computer Vision and Pattern
  Recognition, {CVPR} 2018, Salt Lake City, UT, USA, June 18-22, 2018}, pages
  2724--2732.

\bibitem[{Chang et~al.(2019)Chang, Wang, Peng, and
  Chiu}]{DBLP:conf/cvpr/ChangWPC19}
Wei{-}Lun Chang, Hui{-}Po Wang, Wen{-}Hsiao Peng, and Wei{-}Chen Chiu. 2019.
\newblock \href
  {http://openaccess.thecvf.com/content\_CVPR\_2019/html/Chang\_All\_About\_Structure\_Adapting\_Structural\_Information\_Across\_Domains\_for\_Boosting\_CVPR\_2019\_paper.html}
  {All about structure: Adapting structural information across domains for
  boosting semantic segmentation}.
\newblock In \emph{{IEEE} Conference on Computer Vision and Pattern
  Recognition, {CVPR} 2019, Long Beach, CA, USA, June 16-20, 2019}, pages
  1900--1909.

\bibitem[{Chen et~al.(2011)Chen, Weinberger, and
  Blitzer}]{DBLP:conf/nips/ChenWB11}
Minmin Chen, Kilian~Q. Weinberger, and John Blitzer. 2011.
\newblock \href
  {http://papers.nips.cc/paper/4433-co-training-for-domain-adaptation}
  {Co-training for domain adaptation}.
\newblock In \emph{Advances in Neural Information Processing Systems 24: 25th
  Annual Conference on Neural Information Processing Systems 2011. Proceedings
  of a meeting held 12-14 December 2011, Granada, Spain}, pages 2456--2464.

\bibitem[{Chen et~al.(2012)Chen, Xu, Weinberger, and
  Sha}]{DBLP:conf/icml/ChenXWS12}
Minmin Chen, Zhixiang~Eddie Xu, Kilian~Q. Weinberger, and Fei Sha. 2012.
\newblock \href {http://icml.cc/2012/papers/416.pdf} {Marginalized denoising
  autoencoders for domain adaptation}.
\newblock In \emph{Proceedings of the 29th International Conference on Machine
  Learning, {ICML} 2012, Edinburgh, Scotland, UK, June 26 - July 1, 2012}.

\bibitem[{Dai et~al.(2019)Dai, Shen, Wu, and
  Wang}]{DBLP:journals/corr/abs-1909-01541}
Quanyu Dai, Xiao Shen, Xiao{-}Ming Wu, and Dan Wang. 2019.
\newblock \href {http://arxiv.org/abs/1909.01541} {Network transfer learning
  via adversarial domain adaptation with graph convolution}.
\newblock \emph{CoRR}, abs/1909.01541.

\bibitem[{Daum{\'{e}}~III and Marcu(2006)}]{DBLP:journals/jair/DaumeM06}
Hal Daum{\'{e}}~III and Daniel Marcu. 2006.
\newblock \href {https://doi.org/10.1613/jair.1872} {Domain adaptation for
  statistical classifiers}.
\newblock \emph{J. Artif. Intell. Res.}, 26:101--126.

\bibitem[{Deng et~al.(2018)Deng, Shen, Yang, Li, Du, Fan, and
  Lei}]{DBLP:conf/coling/DengSYLDFL18}
Yang Deng, Ying Shen, Min Yang, Yaliang Li, Nan Du, Wei Fan, and Kai Lei. 2018.
\newblock \href {https://www.aclweb.org/anthology/C18-1279/} {Knowledge as {A}
  bridge: Improving cross-domain answer selection with external knowledge}.
\newblock In \emph{Proceedings of the 27th International Conference on
  Computational Linguistics, {COLING} 2018, Santa Fe, New Mexico, USA, August
  20-26, 2018}, pages 3295--3305.

\bibitem[{Elsahar and Gall{\'e}(2019)}]{elsahar-galle-2019-annotate}
Hady Elsahar and Matthias Gall{\'e}. 2019.
\newblock \href {https://doi.org/10.18653/v1/D19-1222} {To annotate or not?
  predicting performance drop under domain shift}.
\newblock In \emph{Proceedings of the 2019 Conference on Empirical Methods in
  Natural Language Processing and the 9th International Joint Conference on
  Natural Language Processing (EMNLP-IJCNLP)}, pages 2163--2173, Hong Kong,
  China. Association for Computational Linguistics.

\bibitem[{Fellbaum(2010)}]{fellbaum2010wordnet}
Christiane Fellbaum. 2010.
\newblock Wordnet.
\newblock In \emph{Theory and applications of ontology: computer applications},
  pages 231--243. Springer.

\bibitem[{Ganin et~al.(2016)Ganin, Ustinova, Ajakan, Germain, Larochelle,
  Laviolette, March, and Lempitsky}]{JMLR:v17:15-239}
Yaroslav Ganin, Evgeniya Ustinova, Hana Ajakan, Pascal Germain, Hugo
  Larochelle, Fran{\c{c}}ois Laviolette, Mario March, and Victor Lempitsky.
  2016.
\newblock \href {http://jmlr.org/papers/v17/15-239.html} {Domain-adversarial
  training of neural networks}.
\newblock \emph{Journal of Machine Learning Research}, 17(59):1--35.

\bibitem[{Ghosal et~al.(2019)Ghosal, Majumder, Poria, Chhaya, and
  Gelbukh}]{ghosal2019dialoguegcn}
Deepanway Ghosal, Navonil Majumder, Soujanya Poria, Niyati Chhaya, and
  Alexander Gelbukh. 2019.
\newblock Dialoguegcn: A graph convolutional neural network for emotion
  recognition in conversation.
\newblock In \emph{Proceedings of the 2019 Conference on Empirical Methods in
  Natural Language Processing and the 9th International Joint Conference on
  Natural Language Processing (EMNLP-IJCNLP)}, pages 154--164.

\bibitem[{Glorot et~al.(2011)Glorot, Bordes, and
  Bengio}]{DBLP:conf/icml/GlorotBB11}
Xavier Glorot, Antoine Bordes, and Yoshua Bengio. 2011.
\newblock \href {https://icml.cc/2011/papers/342\_icmlpaper.pdf} {Domain
  adaptation for large-scale sentiment classification: {A} deep learning
  approach}.
\newblock In \emph{Proceedings of the 28th International Conference on Machine
  Learning, {ICML} 2011, Bellevue, Washington, USA, June 28 - July 2, 2011},
  pages 513--520.

\bibitem[{Hamilton et~al.(2016{\natexlab{a}})Hamilton, Clark, Leskovec, and
  Jurafsky}]{DBLP:conf/emnlp/HamiltonCLJ16}
William~L. Hamilton, Kevin Clark, Jure Leskovec, and Dan Jurafsky.
  2016{\natexlab{a}}.
\newblock \href {https://www.aclweb.org/anthology/D16-1057/} {Inducing
  domain-specific sentiment lexicons from unlabeled corpora}.
\newblock In \emph{Proceedings of the 2016 Conference on Empirical Methods in
  Natural Language Processing, {EMNLP} 2016, Austin, Texas, USA, November 1-4,
  2016}, pages 595--605. The Association for Computational Linguistics.

\bibitem[{Hamilton et~al.(2016{\natexlab{b}})Hamilton, Leskovec, and
  Jurafsky}]{hamilton-etal-2016-diachronic}
William~L. Hamilton, Jure Leskovec, and Dan Jurafsky. 2016{\natexlab{b}}.
\newblock \href {https://doi.org/10.18653/v1/P16-1141} {Diachronic word
  embeddings reveal statistical laws of semantic change}.
\newblock In \emph{Proceedings of the 54th Annual Meeting of the Association
  for Computational Linguistics (Volume 1: Long Papers)}, pages 1489--1501,
  Berlin, Germany. Association for Computational Linguistics.

\bibitem[{He and Zhou(2011)}]{DBLP:journals/ipm/HeZ11}
Yulan He and Deyu Zhou. 2011.
\newblock \href {https://doi.org/10.1016/j.ipm.2010.11.003} {Self-training from
  labeled features for sentiment analysis}.
\newblock \emph{Inf. Process. Manage.}, 47(4):606--616.

\bibitem[{IV et~al.(2019)IV, Liu, Peters, Gardner, and
  Singh}]{DBLP:conf/acl/LoganLPGS19}
Robert L.~Logan IV, Nelson~F. Liu, Matthew~E. Peters, Matt Gardner, and Sameer
  Singh. 2019.
\newblock \href {https://www.aclweb.org/anthology/P19-1598/} {Barack's wife
  hillary: Using knowledge graphs for fact-aware language modeling}.
\newblock In \emph{Proceedings of the 57th Conference of the Association for
  Computational Linguistics, {ACL} 2019, Florence, Italy, July 28- August 2,
  2019, Volume 1: Long Papers}, pages 5962--5971. Association for Computational
  Linguistics.

\bibitem[{Jiang and Zhai(2007)}]{DBLP:conf/acl/JiangZ07}
Jing Jiang and ChengXiang Zhai. 2007.
\newblock \href {https://www.aclweb.org/anthology/P07-1034/} {Instance
  weighting for domain adaptation in {NLP}}.
\newblock In \emph{{ACL} 2007, Proceedings of the 45th Annual Meeting of the
  Association for Computational Linguistics, June 23-30, 2007, Prague, Czech
  Republic}.

\bibitem[{K~Sarma et~al.(2019)K~Sarma, Liang, and
  Sethares}]{k-sarma-etal-2019-shallow}
Prathusha K~Sarma, Yingyu Liang, and William Sethares. 2019.
\newblock \href {https://doi.org/10.18653/v1/D19-1557} {Shallow domain adaptive
  embeddings for sentiment analysis}.
\newblock In \emph{Proceedings of the 2019 Conference on Empirical Methods in
  Natural Language Processing and the 9th International Joint Conference on
  Natural Language Processing (EMNLP-IJCNLP)}, pages 5548--5557, Hong Kong,
  China. Association for Computational Linguistics.

\bibitem[{Kim et~al.(2017{\natexlab{a}})Kim, Cha, Kim, Lee, and
  Kim}]{DBLP:conf/icml/KimCKLK17}
Taeksoo Kim, Moonsu Cha, Hyunsoo Kim, Jung~Kwon Lee, and Jiwon Kim.
  2017{\natexlab{a}}.
\newblock \href {http://proceedings.mlr.press/v70/kim17a.html} {Learning to
  discover cross-domain relations with generative adversarial networks}.
\newblock In \emph{Proceedings of the 34th International Conference on Machine
  Learning, {ICML} 2017, Sydney, NSW, Australia, 6-11 August 2017}, pages
  1857--1865.

\bibitem[{Kim et~al.(2017{\natexlab{b}})Kim, Stratos, and
  Kim}]{DBLP:conf/acl/KimSK17a}
Young{-}Bum Kim, Karl Stratos, and Dongchan Kim. 2017{\natexlab{b}}.
\newblock \href {https://doi.org/10.18653/v1/P17-1119} {Adversarial adaptation
  of synthetic or stale data}.
\newblock In \emph{Proceedings of the 55th Annual Meeting of the Association
  for Computational Linguistics, {ACL} 2017, Vancouver, Canada, July 30 -
  August 4, Volume 1: Long Papers}, pages 1297--1307.

\bibitem[{Kingma and Ba(2015)}]{DBLP:journals/corr/KingmaB14}
Diederik~P. Kingma and Jimmy Ba. 2015.
\newblock \href {http://arxiv.org/abs/1412.6980} {Adam: {A} method for
  stochastic optimization}.
\newblock In \emph{3rd International Conference on Learning Representations,
  {ICLR} 2015, San Diego, CA, USA, May 7-9, 2015, Conference Track
  Proceedings}.

\bibitem[{Kouw and Loog(2019)}]{kouw2019review}
Wouter~Marco Kouw and Marco Loog. 2019.
\newblock A review of domain adaptation without target labels.
\newblock \emph{IEEE transactions on pattern analysis and machine
  intelligence}.

\bibitem[{Li et~al.(2012)Li, Jin, and Long}]{DBLP:conf/aaai/LiJL12}
Lianghao Li, Xiaoming Jin, and Mingsheng Long. 2012.
\newblock \href {http://www.aaai.org/ocs/index.php/AAAI/AAAI12/paper/view/5077}
  {Topic correlation analysis for cross-domain text classification}.
\newblock In \emph{Proceedings of the Twenty-Sixth {AAAI} Conference on
  Artificial Intelligence, July 22-26, 2012, Toronto, Ontario, Canada}. {AAAI}
  Press.

\bibitem[{Li et~al.(2019)Li, Mao, Yang, and Li}]{li-etal-2019-improving}
Pengfei Li, Kezhi Mao, Xuefeng Yang, and Qi~Li. 2019.
\newblock \href {https://doi.org/10.18653/v1/D19-1022} {Improving relation
  extraction with knowledge-attention}.
\newblock In \emph{Proceedings of the 2019 Conference on Empirical Methods in
  Natural Language Processing and the 9th International Joint Conference on
  Natural Language Processing (EMNLP-IJCNLP)}, pages 229--239, Hong Kong,
  China. Association for Computational Linguistics.

\bibitem[{Li et~al.(2018)Li, Baldwin, and Cohn}]{DBLP:conf/naacl/LiBC18}
Yitong Li, Timothy Baldwin, and Trevor Cohn. 2018.
\newblock \href {https://www.aclweb.org/anthology/N18-2076/} {What's in a
  domain? learning domain-robust text representations using adversarial
  training}.
\newblock In \emph{Proceedings of the 2018 Conference of the North American
  Chapter of the Association for Computational Linguistics: Human Language
  Technologies, NAACL-HLT, New Orleans, Louisiana, USA, June 1-6, 2018, Volume
  2 (Short Papers)}, pages 474--479.

\bibitem[{Liu et~al.(2019)Liu, Long, Wang, and Jordan}]{liu2019transferable}
Hong Liu, Mingsheng Long, Jianmin Wang, and Michael Jordan. 2019.
\newblock Transferable adversarial training: A general approach to adapting
  deep classifiers.
\newblock In \emph{International Conference on Machine Learning}, pages
  4013--4022.

\bibitem[{Liu et~al.(2017)Liu, Qiu, and Huang}]{DBLP:conf/acl/LiuQH17}
Pengfei Liu, Xipeng Qiu, and Xuanjing Huang. 2017.
\newblock \href {https://doi.org/10.18653/v1/P17-1001} {Adversarial multi-task
  learning for text classification}.
\newblock In \emph{Proceedings of the 55th Annual Meeting of the Association
  for Computational Linguistics, {ACL} 2017, Vancouver, Canada, July 30 -
  August 4, Volume 1: Long Papers}, pages 1--10.

\bibitem[{Liu et~al.(2018)Liu, Zhang, and Liu}]{DBLP:conf/naacl/LiuZL18}
Qi~Liu, Yue Zhang, and Jiangming Liu. 2018.
\newblock \href {https://www.aclweb.org/anthology/N18-1050/} {Learning domain
  representation for multi-domain sentiment classification}.
\newblock In \emph{Proceedings of the 2018 Conference of the North American
  Chapter of the Association for Computational Linguistics: Human Language
  Technologies, {NAACL-HLT} 2018, New Orleans, Louisiana, USA, June 1-6, 2018,
  Volume 1 (Long Papers)}, pages 541--550.

\bibitem[{liu et~al.(2019)liu, Niu, Wu, and
  Wang}]{liu-etal-2019-knowledge-aware}
zhibin liu, Zheng-Yu Niu, Hua Wu, and Haifeng Wang. 2019.
\newblock \href {https://doi.org/10.18653/v1/D19-1187} {Knowledge aware
  conversation generation with explainable reasoning over augmented graphs}.
\newblock In \emph{Proceedings of the 2019 Conference on Empirical Methods in
  Natural Language Processing and the 9th International Joint Conference on
  Natural Language Processing (EMNLP-IJCNLP)}, pages 1782--1792, Hong Kong,
  China. Association for Computational Linguistics.

\bibitem[{Louizos et~al.(2015)Louizos, Swersky, Li, Welling, and
  Zemel}]{louizos2015variational}
Christos Louizos, Kevin Swersky, Yujia Li, Max Welling, and Richard Zemel.
  2015.
\newblock The variational fair autoencoder.
\newblock \emph{arXiv preprint arXiv:1511.00830}.

\bibitem[{Ni et~al.(2018)Ni, Chang, Liu, Cheng, Chen, Xu, and
  Zhang}]{DBLP:conf/www/NiCLCCX018}
Jingchao Ni, Shiyu Chang, Xiao Liu, Wei Cheng, Haifeng Chen, Dongkuan Xu, and
  Xiang Zhang. 2018.
\newblock \href {https://doi.org/10.1145/3178876.3186113} {Co-regularized deep
  multi-network embedding}.
\newblock In \emph{Proceedings of the 2018 World Wide Web Conference on World
  Wide Web, {WWW} 2018, Lyon, France, April 23-27, 2018}, pages 469--478.

\bibitem[{Pan et~al.(2010)Pan, Ni, Sun, Yang, and
  Chen}]{DBLP:conf/www/PanNSYC10}
Sinno~Jialin Pan, Xiaochuan Ni, Jian{-}Tao Sun, Qiang Yang, and Zheng Chen.
  2010.
\newblock \href {https://doi.org/10.1145/1772690.1772767} {Cross-domain
  sentiment classification via spectral feature alignment}.
\newblock In \emph{Proceedings of the 19th International Conference on World
  Wide Web, {WWW} 2010, Raleigh, North Carolina, USA, April 26-30, 2010}, pages
  751--760.

\bibitem[{Peng et~al.(2018{\natexlab{a}})Peng, Zhang, Jiang, and
  Huang}]{peng2018cross}
Minlong Peng, Qi~Zhang, Yu-gang Jiang, and Xuan-Jing Huang. 2018{\natexlab{a}}.
\newblock Cross-domain sentiment classification with target domain specific
  information.
\newblock In \emph{Proceedings of the 56th Annual Meeting of the Association
  for Computational Linguistics (Volume 1: Long Papers)}, pages 2505--2513.

\bibitem[{Peng et~al.(2018{\natexlab{b}})Peng, Zhang, Jiang, and
  Huang}]{DBLP:conf/acl/ZhangHPJ18}
Minlong Peng, Qi~Zhang, Yu{-}Gang Jiang, and Xuanjing Huang.
  2018{\natexlab{b}}.
\newblock \href {https://doi.org/10.18653/v1/P18-1233} {Cross-domain sentiment
  classification with target domain specific information}.
\newblock In \emph{Proceedings of the 56th Annual Meeting of the Association
  for Computational Linguistics, {ACL} 2018, Melbourne, Australia, July 15-20,
  2018, Volume 1: Long Papers}, pages 2505--2513.

\bibitem[{Pennington et~al.(2014)Pennington, Socher, and
  Manning}]{DBLP:conf/emnlp/PenningtonSM14}
Jeffrey Pennington, Richard Socher, and Christopher~D. Manning. 2014.
\newblock \href {https://www.aclweb.org/anthology/D14-1162/} {Glove: Global
  vectors for word representation}.
\newblock In \emph{Proceedings of the 2014 Conference on Empirical Methods in
  Natural Language Processing, {EMNLP} 2014, October 25-29, 2014, Doha, Qatar,
  {A} meeting of SIGDAT, a Special Interest Group of the {ACL}}, pages
  1532--1543. {ACL}.

\bibitem[{Peters et~al.(2019)Peters, Neumann, Logan, Schwartz, Joshi, Singh,
  and Smith}]{peters-etal-2019-knowledge}
Matthew~E. Peters, Mark Neumann, Robert Logan, Roy Schwartz, Vidur Joshi,
  Sameer Singh, and Noah~A. Smith. 2019.
\newblock \href {https://doi.org/10.18653/v1/D19-1005} {Knowledge enhanced
  contextual word representations}.
\newblock In \emph{Proceedings of the 2019 Conference on Empirical Methods in
  Natural Language Processing and the 9th International Joint Conference on
  Natural Language Processing (EMNLP-IJCNLP)}, pages 43--54, Hong Kong, China.
  Association for Computational Linguistics.

\bibitem[{Poria et~al.(2020)Poria, Hazarika, Majumder, and
  Mihalcea}]{poria2020beneath}
Soujanya Poria, Devamanyu Hazarika, Navonil Majumder, and Rada Mihalcea. 2020.
\newblock Beneath the tip of the iceberg: Current challenges and new directions
  in sentiment analysis research.
\newblock \emph{arXiv preprint arXiv:2005}.

\bibitem[{Poria et~al.(2019)Poria, Majumder, Mihalcea, and
  Hovy}]{poria2019emotion}
Soujanya Poria, Navonil Majumder, Rada Mihalcea, and Eduard Hovy. 2019.
\newblock Emotion recognition in conversation: Research challenges, datasets,
  and recent advances.
\newblock \emph{IEEE Access}, 7:100943--100953.

\bibitem[{Ruder(2019)}]{ruder2019neural}
Sebastian Ruder. 2019.
\newblock \emph{Neural Transfer Learning for Natural Language Processing}.
\newblock Ph.D. thesis, NATIONAL UNIVERSITY OF IRELAND, GALWAY.

\bibitem[{Ruder and Plank(2018)}]{DBLP:conf/acl/PlankR18}
Sebastian Ruder and Barbara Plank. 2018.
\newblock \href {https://doi.org/10.18653/v1/P18-1096} {Strong baselines for
  neural semi-supervised learning under domain shift}.
\newblock In \emph{Proceedings of the 56th Annual Meeting of the Association
  for Computational Linguistics, {ACL} 2018, Melbourne, Australia, July 15-20,
  2018, Volume 1: Long Papers}, pages 1044--1054. Association for Computational
  Linguistics.

\bibitem[{Saito et~al.(2017)Saito, Ushiku, and Harada}]{saito2017asymmetric}
Kuniaki Saito, Yoshitaka Ushiku, and Tatsuya Harada. 2017.
\newblock Asymmetric tri-training for unsupervised domain adaptation.
\newblock In \emph{Proceedings of the 34th International Conference on Machine
  Learning-Volume 70}, pages 2988--2997. JMLR. org.

\bibitem[{Sarma et~al.(2018)Sarma, Liang, and
  Sethares}]{DBLP:conf/acl/SarmaLS18}
Prathusha~K. Sarma, Yingyu Liang, and Bill Sethares. 2018.
\newblock \href {https://doi.org/10.18653/v1/P18-2007} {Domain adapted word
  embeddings for improved sentiment classification}.
\newblock In \emph{Proceedings of the 56th Annual Meeting of the Association
  for Computational Linguistics, {ACL} 2018, Melbourne, Australia, July 15-20,
  2018, Volume 2: Short Papers}, pages 37--42.

\bibitem[{Schlichtkrull et~al.(2018)Schlichtkrull, Kipf, Bloem, van~den Berg,
  Titov, and Welling}]{DBLP:conf/esws/SchlichtkrullKB18}
Michael~Sejr Schlichtkrull, Thomas~N. Kipf, Peter Bloem, Rianne van~den Berg,
  Ivan Titov, and Max Welling. 2018.
\newblock \href {https://doi.org/10.1007/978-3-319-93417-4\_38} {Modeling
  relational data with graph convolutional networks}.
\newblock In \emph{The Semantic Web - 15th International Conference, {ESWC}
  2018, Heraklion, Crete, Greece, June 3-7, 2018, Proceedings}, volume 10843 of
  \emph{Lecture Notes in Computer Science}, pages 593--607. Springer.

\bibitem[{Sharma et~al.(2018)Sharma, Bhattacharyya, Dandapat, and
  Bhatt}]{DBLP:conf/acl/BhattacharyyaDS18}
Raksha Sharma, Pushpak Bhattacharyya, Sandipan Dandapat, and Himanshu~Sharad
  Bhatt. 2018.
\newblock \href {https://doi.org/10.18653/v1/P18-1089} {Identifying
  transferable information across domains for cross-domain sentiment
  classification}.
\newblock In \emph{Proceedings of the 56th Annual Meeting of the Association
  for Computational Linguistics, {ACL} 2018, Melbourne, Australia, July 15-20,
  2018, Volume 1: Long Papers}, pages 968--978.

\bibitem[{Shen and Chung(2019)}]{DBLP:journals/corr/abs-1901-07264}
Xiao Shen and Fu{-}Lai Chung. 2019.
\newblock \href {http://arxiv.org/abs/1901.07264} {Network embedding for
  cross-network node classification}.
\newblock \emph{CoRR}, abs/1901.07264.

\bibitem[{Shi et~al.(2018)Shi, Fu, Bing, and Lam}]{DBLP:conf/acl/LamSBF18}
Bei Shi, Zihao Fu, Lidong Bing, and Wai Lam. 2018.
\newblock \href {https://doi.org/10.18653/v1/P18-1232} {Learning
  domain-sensitive and sentiment-aware word embeddings}.
\newblock In \emph{Proceedings of the 56th Annual Meeting of the Association
  for Computational Linguistics, {ACL} 2018, Melbourne, Australia, July 15-20,
  2018, Volume 1: Long Papers}, pages 2494--2504. Association for Computational
  Linguistics.

\bibitem[{Speer et~al.(2017)Speer, Chin, and Havasi}]{speer2017conceptnet}
Robert Speer, Joshua Chin, and Catherine Havasi. 2017.
\newblock Conceptnet 5.5: An open multilingual graph of general knowledge.
\newblock In \emph{Thirty-First AAAI Conference on Artificial Intelligence}.

\bibitem[{Tzeng et~al.(2017)Tzeng, Hoffman, Saenko, and
  Darrell}]{DBLP:conf/cvpr/TzengHSD17}
Eric Tzeng, Judy Hoffman, Kate Saenko, and Trevor Darrell. 2017.
\newblock \href {https://doi.org/10.1109/CVPR.2017.316} {Adversarial
  discriminative domain adaptation}.
\newblock In \emph{2017 {IEEE} Conference on Computer Vision and Pattern
  Recognition, {CVPR} 2017, Honolulu, HI, USA, July 21-26, 2017}, pages
  2962--2971. {IEEE} Computer Society.

\bibitem[{Tzeng et~al.(2014)Tzeng, Hoffman, Zhang, Saenko, and
  Darrell}]{DBLP:journals/corr/TzengHZSD14}
Eric Tzeng, Judy Hoffman, Ning Zhang, Kate Saenko, and Trevor Darrell. 2014.
\newblock \href {http://arxiv.org/abs/1412.3474} {Deep domain confusion:
  Maximizing for domain invariance}.
\newblock \emph{CoRR}, abs/1412.3474.

\bibitem[{Wang et~al.(2008)Wang, Domeniconi, and Hu}]{DBLP:conf/icdm/WangDH08}
Pu~Wang, Carlotta Domeniconi, and Jian Hu. 2008.
\newblock \href {https://doi.org/10.1109/ICDM.2008.136} {Using wikipedia for
  co-clustering based cross-domain text classification}.
\newblock In \emph{Proceedings of the 8th {IEEE} International Conference on
  Data Mining {(ICDM} 2008), December 15-19, 2008, Pisa, Italy}, pages
  1085--1090. {IEEE} Computer Society.

\bibitem[{Wilson and Cook(2018)}]{DBLP:journals/corr/abs-1812-02849}
Garrett Wilson and Diane~J. Cook. 2018.
\newblock \href {http://arxiv.org/abs/1812.02849} {Adversarial transfer
  learning}.
\newblock \emph{CoRR}, abs/1812.02849.

\bibitem[{Xiang et~al.(2010)Xiang, Cao, Hu, and
  Yang}]{DBLP:journals/tkde/XiangCHY10}
Evan~Wei Xiang, Bin Cao, Derek~Hao Hu, and Qiang Yang. 2010.
\newblock \href {https://doi.org/10.1109/TKDE.2010.31} {Bridging domains using
  world wide knowledge for transfer learning}.
\newblock \emph{{IEEE} Trans. Knowl. Data Eng.}, 22(6):770--783.

\bibitem[{Xu et~al.(2017)Xu, Wei, Cao, and Yu}]{DBLP:conf/wsdm/XuWCY17}
Linchuan Xu, Xiaokai Wei, Jiannong Cao, and Philip~S. Yu. 2017.
\newblock \href {https://doi.org/10.1145/3018661.3018723} {Embedding of
  embedding {(EOE):} joint embedding for coupled heterogeneous networks}.
\newblock In \emph{Proceedings of the Tenth {ACM} International Conference on
  Web Search and Data Mining, {WSDM} 2017, Cambridge, United Kingdom, February
  6-10, 2017}, pages 741--749. {ACM}.

\bibitem[{Yang et~al.(2014)Yang, Yih, He, Gao, and Deng}]{yang2014embedding}
Bishan Yang, Wen-tau Yih, Xiaodong He, Jianfeng Gao, and Li~Deng. 2014.
\newblock Embedding entities and relations for learning and inference in
  knowledge bases.
\newblock \emph{arXiv preprint arXiv:1412.6575}.

\bibitem[{Yang et~al.(2019)Yang, Li, Luo, Liu, and
  Sun}]{yang-etal-2019-enhancing-topic}
Pengcheng Yang, Lei Li, Fuli Luo, Tianyu Liu, and Xu~Sun. 2019.
\newblock \href {https://doi.org/10.18653/v1/P19-1193} {Enhancing
  topic-to-essay generation with external commonsense knowledge}.
\newblock In \emph{Proceedings of the 57th Annual Meeting of the Association
  for Computational Linguistics}, pages 2002--2012, Florence, Italy.
  Association for Computational Linguistics.

\bibitem[{Zellinger et~al.(2017)Zellinger, Grubinger, Lughofer,
  Natschl{\"a}ger, and Saminger-Platz}]{zellinger2017central}
Werner Zellinger, Thomas Grubinger, Edwin Lughofer, Thomas Natschl{\"a}ger, and
  Susanne Saminger-Platz. 2017.
\newblock Central moment discrepancy (cmd) for domain-invariant representation
  learning.
\newblock \emph{arXiv preprint arXiv:1702.08811}.

\bibitem[{Zhang et~al.(2018)Zhang, Wang, and
  Liu}]{DBLP:journals/widm/ZhangWL18}
Lei Zhang, Shuai Wang, and Bing Liu. 2018.
\newblock \href {https://doi.org/10.1002/widm.1253} {Deep learning for
  sentiment analysis: {A} survey}.
\newblock \emph{Wiley Interdiscip. Rev. Data Min. Knowl. Discov.}, 8(4).

\bibitem[{Zhong et~al.(2019)Zhong, Wang, and Miao}]{zhong-etal-2019-knowledge}
Peixiang Zhong, Di~Wang, and Chunyan Miao. 2019.
\newblock \href {https://doi.org/10.18653/v1/D19-1016} {Knowledge-enriched
  transformer for emotion detection in textual conversations}.
\newblock In \emph{Proceedings of the 2019 Conference on Empirical Methods in
  Natural Language Processing and the 9th International Joint Conference on
  Natural Language Processing (EMNLP-IJCNLP)}, pages 165--176, Hong Kong,
  China. Association for Computational Linguistics.

\bibitem[{Ziser and Reichart(2019)}]{ziser2019task}
Yftah Ziser and Roi Reichart. 2019.
\newblock Task refinement learning for improved accuracy and stability of
  unsupervised domain adaptation.
\newblock In \emph{Proceedings of the 57th Annual Meeting of the Association
  for Computational Linguistics}, pages 5895--5906.

\end{thebibliography}
\bibliographystyle{acl_natbib}


\end{document}